\pdfoutput=1

\documentclass[11pt]{article}

\usepackage[final]{acl}

\usepackage{amsmath,amsfonts,bm}

\def\eqref#1{equation~\ref{#1}}

\def\1{\bm{1}}

\DeclareMathAlphabet{\mathsfit}{\encodingdefault}{\sfdefault}{m}{sl}
\SetMathAlphabet{\mathsfit}{bold}{\encodingdefault}{\sfdefault}{bx}{n}

\PassOptionsToPackage{hyphens}{url}
\usepackage{hyperref}
\hypersetup{
    colorlinks=true,       %
    linkcolor=red,        %
    citecolor=blue,         %
    filecolor=magenta,     %
    urlcolor=cyan          %
}
\usepackage{multirow}
\usepackage[shortlabels]{enumitem}
\usepackage{url}
\usepackage[utf8]{inputenc} %

\usepackage[T1]{fontenc}    %
\usepackage{booktabs,stmaryrd}       %
\usepackage{nicefrac}       %
\usepackage{microtype}      %
\usepackage{xcolor}         %
\usepackage{makecell}
\usepackage{subfig}

\usepackage{amssymb}
\usepackage{amsmath,amsthm,amsfonts,bm}
\usepackage{graphicx}
\usepackage[noend]{algorithmic}
\usepackage[algoruled,nofillcomment,algo2e]{algorithm2e}
\usepackage{placeins}
\usepackage{array}
\usepackage{dsfont}
\usepackage{bbold}
\usepackage{tikz}
\usepackage{comment}
\usetikzlibrary{positioning}
\usepackage{float} 
\usepackage{colortbl}

\newcommand{\myparagraph}[1]{\vspace{2pt}\noindent{\bf #1}}

\def\name{\textit{CAMIA}\xspace}

\usepackage[colorinlistoftodos,textsize=small]{todonotes}

\usepackage{times}
\usepackage{latexsym}

\usepackage[T1]{fontenc}

\usepackage[utf8]{inputenc}

\usepackage{microtype}

\usepackage{inconsolata}

\usepackage{graphicx}
\theoremstyle{definition}

\usepackage{color}
\usepackage{tabularx, booktabs}
\usepackage{tabularx, booktabs}
\newcolumntype{Y}{>{\centering\arraybackslash}X}

\usepackage{xcolor,colortbl} %

\title{Context-Aware Membership Inference Attacks\\ against Pre-trained Large Language Models}
\author{
  \textbf{Hongyan Chang}\textsuperscript{1}\thanks{The work was completed during Hongyan's internship at Brave.} \quad
  \textbf{Ali Shahin Shamsabadi}\textsuperscript{2}\quad
  \textbf{Kleomenis Katevas}\textsuperscript{2} \\
  \textbf{Hamed Haddadi}\textsuperscript{2,3} \quad
  \textbf{Reza Shokri}\textsuperscript{4} \\[1em]
  \textsuperscript{1}Mohamed bin Zayed University of Artificial Intelligence \quad
  \textsuperscript{2}Brave Software \\
  \textsuperscript{3}Imperial College London \quad
  \textsuperscript{4}National University of Singapore \\[0.5em]
}
\begin{document}

\maketitle
\begin{abstract}
Membership Inference Attacks (MIAs) on pre-trained Large Language Models (LLMs) aim at determining if a data point was part of the model's training set. Prior MIAs that are built for classification models fail at LLMs, due to ignoring the generative nature of LLMs across token sequences. In this paper, we present a novel attack on pre-trained LLMs that adapts MIA statistical tests to the perplexity dynamics of subsequences within a data point. Our method significantly outperforms prior approaches, revealing context-dependent memorization patterns in pre-trained LLMs.
\end{abstract}

\section{Introduction}\label{sec:intro}

To assess memorization and information leakage in models, Membership Inference Attacks (MIAs) aim to determine if a data point was part of a model's training set~\cite{shokri2017membership}. However, MIAs designed for pre-trained Large Language Models (LLMs) have been largely ineffective~\cite{duan2024membership,das2024blind}.

This is primarily because these MIAs, originally developed for classification models, fail to account for the generative nature of LLMs. Unlike classification models, which produce a single prediction based on the input, LLMs generate texts token-by-token, adjusting the prediction for each output token based on the \textit{context} of preceding tokens (i.e., the prefix). Prior MIAs overlook this \textit{token-level loss dynamics} and the \textit{influence of prefixes} on the predicted token, both of which contribute to the memorization behaviors of LLMs. As such simplifications miss the critical behaviors of LLMs, notably \textit{context-dependent memorization}, these attacks are often ineffective at identifying training set members in pre-trained LLMs. 

Additionally, state-of-the-art MIAs~\cite{zarifzadeh2024low, carlini2022membership, ye2022enhanced, mireshghallah2022quantifying} rely on reference models trained similarly to the target model but on a distinct but similarly distributed dataset. Obtaining such reference models is extremely costly and often impractical for pre-trained LLMs. On the other hand, using other available pre-trained models as reference models may also lead to inaccurate attacks due to significant differences in their training processes and model architectures (as is shown analytically~\cite{murakonda2021quantifying} and empirically~\cite{duan2024membership} in the literature). 

To design a strong MIA against pre-trained LLMs, we need to fully understand how and why memorization occurs during the training. Any piece of text is modeled as a sequence of tokens, and LLMs are trained to maximize the conditional probabilities of generating each token based on the preceding context (i.e., the prefix), by adjusting the model parameters. This process is progressive, as the model adjusts its predictions with each new token, refining its understanding of the sequence. 

\vspace{2pt}
\noindent\textbf{Key attack insight.} Our insight is that memorization is context-dependent, triggered primarily when the prefix provides insufficient information for accurate next-token prediction. If a prefix clearly constrains the possible next tokens, either because it contains repetitive patterns or the next tokens overlap strongly with prefix content, the model can reliably predict the next token through generalization, without significant memorization. In contrast, when the prefix is ambiguous or complex, failing to clearly narrow down subsequent possibilities, the model becomes uncertain. To resolve this uncertainty, the model is more likely to rely on specific memorized sequences encountered during training. Therefore, rather than simply relying on the overall loss across a text sequence as in prior work, an effective MIA must account explicitly for how context influences the model's predictive uncertainty at the token level.

Motivated by this insight, we propose \name, a Context-Aware Membership Inference Attack specifically designed to exploit the relationship between prefix ambiguity and memorization. The core idea behind \name is straightforward yet powerful: it analyzes how quickly and stably the model transitions from initial uncertainty (high ambiguity) to confident predictions as it generates tokens. By capturing the rate at which prediction uncertainty is resolved, as well as correcting for scenarios where ambiguity is artificially reduced by repetitive content, our method effectively distinguishes memorized sequences from generalized predictions. Unlike prior attacks that rely on static thresholds for average prediction losses, \name dynamically adapts its inference strategy at the token-level, directly leveraging the context-dependent nature of memorization to significantly improve inference accuracy.

We provide a comprehensive evaluation of our \name on a wide spectrum of pre-trained LLMs from the Pythia~\cite{biderman2023pythia} and GPT-Neo~\cite{gpt-neo} suites against prior attacks on the MIMIR benchmark~\cite{duan2024membership}. The performance increase of our attack is consistent across models of various sizes and $6$ data domains. For instance, when attacking the $2.8$B Pythia model on member/non-member data sampled from the Arxiv domain, \name successfully identifies almost twice more members than prior baselines, increasing the true positive rate from $20.11\%$ to $32\%$ while maintaining a $1\%$ false-positive error rate.
\footnote{The code is available in \url{https://github.com/changhongyan123/context_aware_mia}} 

\section{Problem Formulation}
\label{sec:problem_formulation}

\subsection{Autoregressive language model training}
Let $\mathcal{M}$ be an auto-regressive model trained on a private dataset $\text{PrivSet} = \{\mathbf{X}_i\}_{i=1}^N$ of size $N$. Each text $\mathbf{X}_i$ is tokenized into $T$ tokens via a token embedding function, forming a sequence $\{x_1, \ldots, x_T\}$ over a vocabulary $\mathbf{V}$. Let $\mathbf{x}_{<t} = \{x_1,\ldots,x_{t-1}\}$ be the prefix of length $t{-}1$. The model $\mathcal{M}$ predicts $x_t$ conditioned on $\mathbf{x}_{<t}$, and the prediction loss is defined as the cross-entropy between the predicted distribution $P(x|\mathbf{x}_{<t};\mathcal{M})$ and the true next token $x_t$:
\begin{align}
\label{eq:tokenLoss}
\mathcal{L}_t(x_t) 
&= -\log P(x_t|\mathbf{x}_{<t};\mathcal{M}).
\end{align}
The model minimizes the average next-token loss: $-\frac{1}{T} \sum_{t=1}^T \mathcal{L}_t(x_t)$.

\subsection{Membership inference attack (MIA)}
\label{sec:mia}
\label{sec:prior_attacks}
MIAs aim to determine whether a target data point $\mathbf{X}$ was part of the training set $\text{PrivSet}$ of a model $\mathcal{M}$. MIAs can be formulated as hypothesis tests: the null hypothesis assumes $\mathbf{X}$ is a non-member, while the alternative assumes it is a member. The adversary’s goal is to decide between the two, incurring false positives (non-members misclassified as members) and false negatives (members misclassified as non-members). Following prior work~\cite{yeom2018privacy,shi2023detecting,zhang2024min,carlini2021extracting}, MIAs typically define a membership score $f(\mathbf{X};\mathcal{M})$ and compare it to a threshold $\tau$ to determine membership.

\myparagraph{Average loss.} A basic approach computes the average next-token loss, $-\frac{1}{T} \sum_{t=1}^T \mathcal{L}_t(x_t)$, and classifies $\mathbf{X}$ as a member if the score is below $\tau$~\cite{yeom2018privacy}.

\myparagraph{Outlier token loss.} Min-K\%~\cite{shi2023detecting} averages the losses over the $k\%$ least likely tokens (i.e., with highest $\mathcal{L}_t$), under the intuition that non-members contain more high-loss outliers. Min-K\%++~\cite{zhang2024min} normalizes each selected token’s loss using the expectation and variance of log-probabilities at its position.

\myparagraph{Loss calibration.} Zlib~\cite{carlini2021extracting} calibrates the loss by dividing by the input’s zlib entropy~\cite{deutsch1996zlib}, i.e., $\mathcal{L}(\mathbf{X};\mathcal{M}) / \text{zlib}(\mathbf{X})$. Reference-based MIA~\cite{carlini2021extracting} compares losses from $\mathcal{M}$ and a reference model $\mathcal{M}_{\text{ref}}$ via $\mathcal{L}(\mathbf{X};\mathcal{M}) - \mathcal{L}(\mathbf{X};\mathcal{M}_{\text{ref}})$, aiming to isolate training-specific memorization. Neighborhood MIA~\cite{mattern2023membership} subtracts the average loss over neighbors $\mathcal{N}(\mathbf{X})$ from the loss on $\mathbf{X}$:
$
{\mathcal{L}(\mathbf{X};\mathcal{M}) - \frac{1}{|\mathcal{N}(\mathbf{X})|} \sum_{\tilde{\mathbf{X}} \in \mathcal{N}(\mathbf{X})} \mathcal{L}(\tilde{\mathbf{X}}; \mathcal{M}).}
$

\subsection{True leakage vs. MIA effectiveness}
\label{sec:evaluation_mia}

The effectiveness of Membership Inference Attacks (MIAs) depends primarily on two factors: the model's true leakage (memorization tendency) and the design of the attack algorithm. Models with limited memorization behave similarly for members and non-members, making MIAs inherently challenging~\cite{ye2022enhanced,carlini2022membership}. Attack design also critically influences performance; poorly optimized attacks may inaccurately infer membership from prediction correctness alone, failing to account for data and model-specific nuances~\cite{yeom2018privacy,carlini2022membership}. Such challenges motivate our context-aware MIA approach, which explicitly considers context-dependent model behaviors.

Moreover, empirical evaluations of MIAs must carefully handle textual overlaps and dataset construction. Defining membership clearly is particularly difficult for textual data, as minor differences (e.g., punctuation) can obscure exact matches, and overlapping substrings create ambiguity between members and non-members~\cite{duan2024membership}. Benchmarks that artificially distinguish members from non-members based on external factors can inflate measured attack performance. E.g., WikiMIA separates groups by publication date~\cite{shi2023detecting} and a \textit{blind baseline}—which predicts membership solely from text content without any model access—already achieves 98.7\% AUC, indicating that such benchmarks may not accurately reflect genuine model memorization. \textit{Commercial LLMs} such as GPT-4~\cite{achiam2023gpt} often do not disclose training datasets, complicating rigorous evaluations of memorization. To address these evaluation challenges, we adopt the carefully designed MIMIR benchmark~\cite{duan2024membership} and focus our experiments on open-source models (e.g., Pythia~\cite{biderman2023pythia} and GPT-Neo~\cite{gpt-neo}), ensuring transparent and accurate assessment of MIA effectiveness.

\subsection{Threat model and our goal}
\label{sec:our_goal}
We adopt the practical threat model from prior work~\cite{shi2023detecting,zhang2024min}, where the adversary queries the target LLM $\mathcal{M}$ with arbitrary token sequences and obtains per-token losses (Equation~\ref{eq:tokenLoss}), without direct access to $\mathcal{M}$'s architecture or parameters. Practically, per-token loss information was directly accessible via certain APIs (e.g., OpenAI's API before Oct. 2023). Even without direct per-token access, token-level losses can be derived from total sequence losses by comparing incremental predictions (e.g., querying losses for \texttt{"The"} and \texttt{"The sky"})~\cite{yeom2018privacy,carlini2021extracting}.

Under this practical threat model, our primary goal is to enhance MIAs specifically by leveraging the observation that LLMs memorize training data in a context-dependent manner—memorization is particularly pronounced when the prefix does not sufficiently constrain possible next tokens. Prior MIAs typically aggregate losses at the sequence level and thus fail to exploit these finer-grained, token-level memorization patterns. 

To address this limitation, we propose \name, a context-aware MIA framework that explicitly captures how prefixes influence the model's reliance on memorization. Specifically, \name (1) computes per-token prediction losses; (2) extracts signals reflecting context-dependent memorization from these losses; (3) calibrates and combines these signals into tailored membership inference tests; and (4) determines membership status based on these test outcomes. Sections~\ref{sec:bag_of_signal} and~\ref{sec:combine_signal} detail our signal extraction methods and their integration into \name.

\section{Context Aware Membership Signals}
\label{sec:bag_of_signal} 

\subsection{Intuitions}\label{sec:signal_intuition}
As discussed earlier, existing MIAs rely primarily on sequence-level losses, overlooking crucial token-level dynamics driving context-dependent memorization. Below, we clearly illustrate why explicitly modeling token-level prediction difficulty significantly improves MIAs for LLMs.

Traditional MIAs compare input losses to fixed thresholds, ignoring that prediction difficulty varies across data points. Easy-to-predict non-members naturally yield low losses (leading to false positives), while challenging-to-predict members often yield high losses (causing false negatives)~\cite{ye2022enhanced,carlini2022membership}. Prior works address this by calibrating membership inference scores to account explicitly for input-specific difficulty~\cite{zhang2021characterizing,carlini2022membership}. However, existing calibration methods typically rely on computationally expensive approaches (e.g., training the same models without the target sample) or simplistic input-level proxies (e.g., compression-based metrics~\cite{carlini2023extracting}), neglecting the sequential, token-level prediction behavior inherent to LLMs.

In contrast, we focus explicitly on token-level, context-dependent memorization. Specifically, LLMs predict each token sequentially based on preceding tokens (prefix contexts). Memorization emerges most strongly when the prefix provides insufficient predictive guidance, prompting models to rely heavily on memorized training data to resolve uncertainty. For example, consider predicting tokens in the sentence: \emph{``The important thing is not to stop questioning. Curiosity has its own reason for existing.''} Predicting the second occurrence of ``is'' is straightforward regardless of membership status, while predicting ``Curiosity''—due to its ambiguous context—is substantially easier if the model memorized this sentence during training. Such token-level distinctions highlight the importance of explicitly modeling context-dependent memorization for effective MIAs.

\subsection{Signal design}\label{sec:bag_of_signal}
Motivated by this insight, our approach calibrates membership inference directly at the token prediction level. We propose signals specifically designed to capture these subtle, context-dependent memorization patterns—previously overlooked by existing attacks~\cite{carlini2021extracting,shi2023detecting}—as detailed next.

\begin{figure}[t!]
    \centering
\includegraphics[width=\columnwidth]{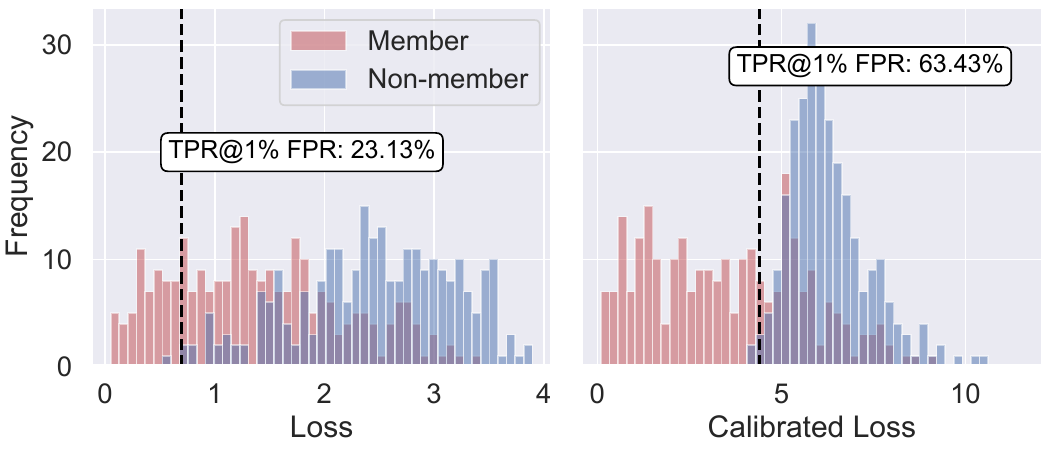}
    \caption{Effect of token diversity calibration on membership inference (Pythia-160M, GitHub domain). Calibration clearly improves separation of members and non-members, significantly enhancing true positive rates (TPR) at low false positive rates (FPR).}
    \label{fig:impact_of_calivration}
\end{figure}

\myparagraph{Token diversity calibration.}
Texts containing repetitive patterns yield inherently lower losses regardless of memorization status~\cite{holtzmancurious,welleckneural}. 
For example, the text \emph{``The cat sat on the mat. The cat sat on the mat.''} naturally produces low loss due to repetition, potentially causing false positives (i.e., predicting a non-member as a member).

To address this bias, we introduce a lightweight calibration based on token diversity:
\begin{align}
   \hspace{-0.6em} d_{\mathbf{X}} = \frac{|Dedup(\mathbf{X})|}{|\mathbf{X}|},\hspace{0.6em}
    f_{\text{Cal}}(\mathbf{X})=\frac{\mathcal{L}(\mathbf{X};\mathcal{M})}{d_{\mathbf{X}}},
\end{align}
where $|Dedup(\mathbf{X})|$ counts unique tokens. 
As shown in Figure~\ref{fig:impact_of_calivration}, this calibration better distinguishes genuinely memorized sequences from trivially predictable repetitive texts.

\myparagraph{Token diversity calibration.}
Repetitive texts naturally yield low losses regardless of memorization, which can cause false positives~\cite{holtzmancurious,welleckneural}. For example, the sequence \emph{``The cat sat on the mat. The cat sat on the mat.''} produces low loss purely due to redundancy. 

To mitigate this, we calibrate losses by token diversity:
\[
   d_{\mathbf{X}} = \frac{|Dedup(\mathbf{X})|}{|\mathbf{X}|},\quad
   f_{\text{Cal}}(\mathbf{X})=\frac{\mathcal{L}(\mathbf{X};\mathcal{M})}{d_{\mathbf{X}}},
\]
where $|Dedup(\mathbf{X})|$ counts unique tokens. As shown in Figure~\ref{fig:impact_of_calivration}, this calibration improves separability between members and non-members in the GitHub domain, where repeated code patterns are common.

\begin{figure}[t!]
\centering
\includegraphics[width=0.98\columnwidth]{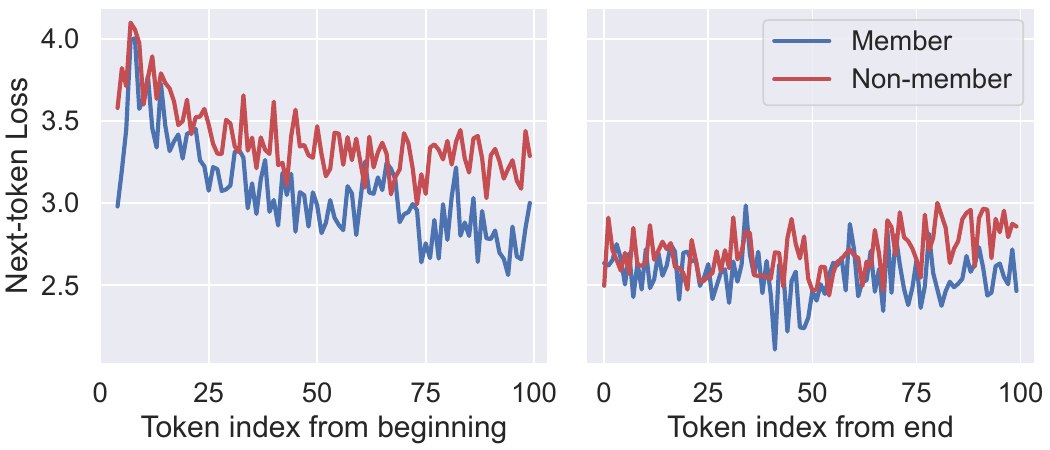}
\caption{Average token losses at the beginning (left) and end (right) of sequences (Pythia-160M, Arxiv domain). Early tokens exhibit clearer differences between members and non-members, justifying our cut-off approach.}
\label{fig:loss_limited_arxiv_160m}
\end{figure}
\myparagraph{Filtering less informative tokens (cut-off loss).}
As prefixes grow longer, contextual cues reduce ambiguity and diminish memorization signals~\cite{levy2024same}. For example, in Figure~\ref{fig:loss_limited_arxiv_160m}, the first few tokens predicted with little context show clear loss gaps between members and non-members, whereas later tokens converge as the prefix becomes increasingly informative. 

To exploit this effect, we truncate the loss sequence to the first $T'$ tokens and leverage the new membership signal ${f_{\text{Cut}}(\mathbf{X}) = \frac{1}{T'}\sum_{t=1}^{T'} \mathcal{L}_t(x_t)}$.

\begin{figure}[t!]
\centering
\includegraphics[width=\columnwidth]{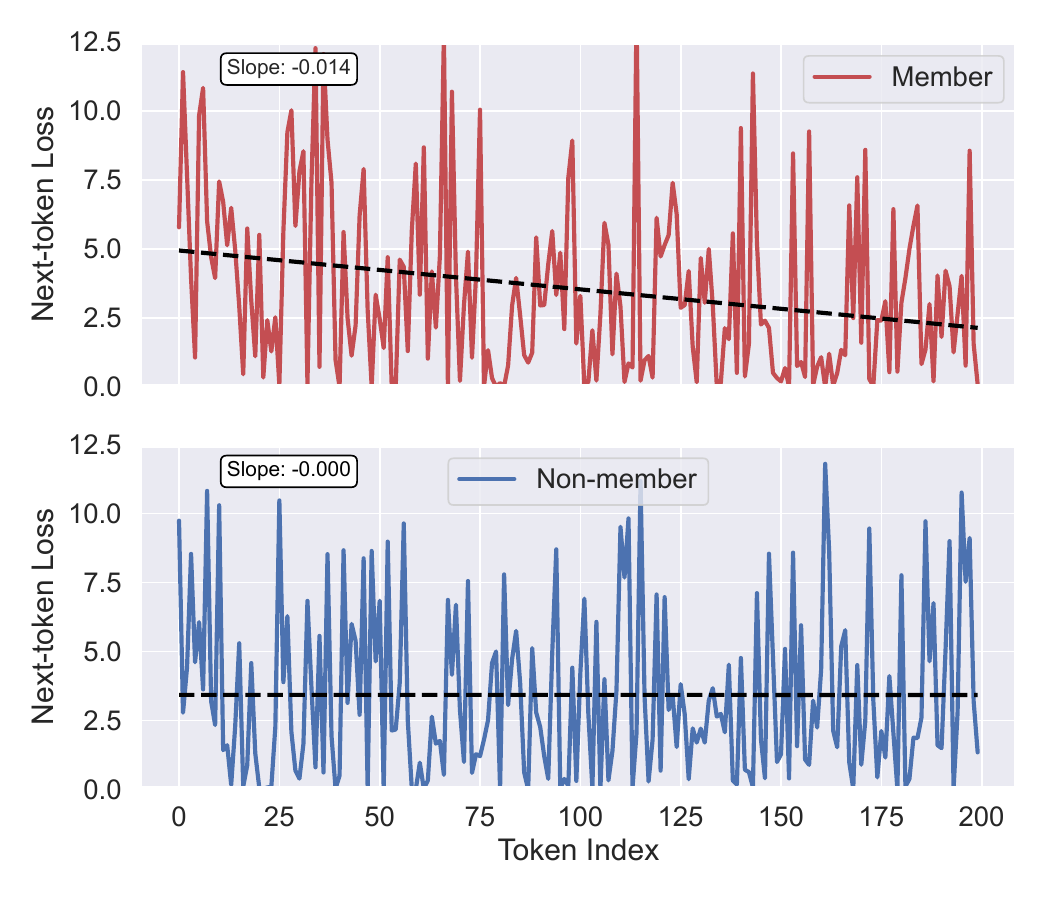}
\caption{Linear fits to token-loss sequences (Pythia-160M, Pile-CC domain). Member losses decrease significantly faster (slope $=0.014$) than non-members (slope $=0.000$), reflecting stronger memorization.}
\label{fig:loss_dynamic_two_example_2.8_pile_cc}
\end{figure}
\myparagraph{Loss decreasing rate (slope).}
When the prefix is ambiguous, memorized continuations quickly reduce uncertainty, leading to faster decreases in token losses. In other words, if the model has encountered the sequence during training, it can immediately resolve the ambiguity and drive losses down, whereas for non-members the model must rely on gradually accumulating context, resulting in a slower decline. For example, Figure~\ref{fig:loss_dynamic_two_example_2.8_pile_cc} shows that member losses decline much more steeply than non-members. 

We capture this effect by fitting a simple linear trend to the first $T'$ token losses, where the slope serves as the signal:
\[
f_{\text{Slope}}(\mathbf{X}) = 
\frac{\sum_{t=1}^{T'} (t-\bar{t}) \big(\mathcal{L}_t(x_t)-\bar{\mathcal{L}}\big)}
     {\sum_{t=1}^{T'} (t-\bar{t})^2},
\]
with $\bar{t}=\frac{T'+1}{2}$ and $\bar{\mathcal{L}}=\tfrac{1}{T'}\sum_{t=1}^{T'} \mathcal{L}_t(x_t)$.

\begin{figure}[t!]
\centering
\includegraphics[width=0.98\columnwidth]{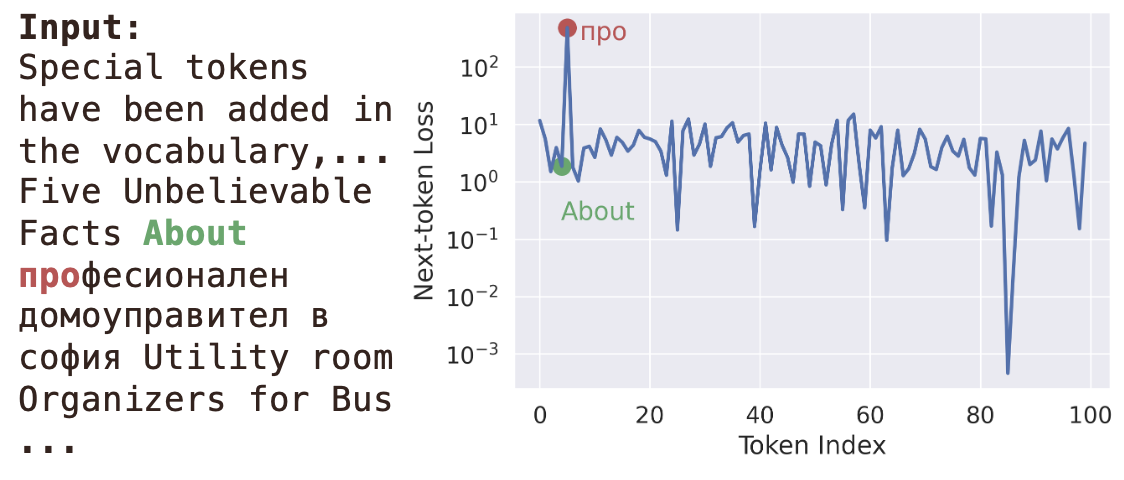}
\caption{Example token-loss spike when the language switches from English to Bulgarian (Pythia-160M). The sudden spike (loss = $484.2$) dominates the sequence, inflating the average loss to $9.33$ and obscuring membership signals. This motivates using robust counts of low-loss tokens rather than relying on average loss.}
\label{fig:loss_increase_change_language}
\end{figure}
\myparagraph{Robust low-loss counting.}
Average losses can be distorted by occasional spikes, for example when the input language suddenly shifts and produces extremely high token losses (Figure~\ref{fig:loss_increase_change_language}). A few ``outlier'' tokens result in a higher average loss value even for members, leading to a false negative error (i.e., predicting a member as a non-member).

To reduce this sensitivity, we instead count how many tokens fall below adaptive loss thresholds, thereby capturing the persistence of low-loss predictions that indicate memorization. We consider three variations: $ f_{\text{CB}}(\mathbf{X}) = \tfrac{1}{T'} \sum_{t=1}^{T'} \mathbb{1}\!\left[\mathcal{L}_t(x_t) \leq \tau \right], $  
which uses a fixed global threshold $\tau$ to measure the overall prevalence of low-loss tokens.  $ f_{\text{CBM}}(\mathbf{X}) = \tfrac{1}{T'} \sum_{t=1}^{T'} \mathbb{1}\!\left[\mathcal{L}_t(x_t) \leq \bar{\mathcal{L}}_{\mathbf{X}} \right], $  
which adapts the threshold to the sequence-level mean $\bar{\mathcal{L}}_{\mathbf{X}}$, normalizing for difficulty across inputs. $ f_{\text{CBPM}}(\mathbf{X}) = \tfrac{1}{T'} \sum_{t=1}^{T'} \mathbb{1}\!\left[\mathcal{L}_t(x_t) \leq \bar{\mathcal{L}}_{\mathbf{X}_{< t}} \right], $  
which uses the running mean loss $\bar{\mathcal{L}}_{\mathbf{X}_{< t}}$ to capture token-level deviations relative to prior context. Together, these variations provide complementary ways of quantifying robust token-level evidence of memorization, less affected by extreme losses.

\myparagraph{Loss fluctuation metrics.}
Non-members often exhibit unstable token-level predictions, reflecting unresolved uncertainty, whereas memorized sequences yield smoother and more regular loss patterns (Figure~\ref{fig:loss_dynamic_two_example_2.8_pile_cc}). Intuitively, when a model recalls training data it can make consistently confident predictions across consecutive tokens, while for unseen inputs its uncertainty fluctuates from token to token. 

To quantify this distinction, we employ two sequence-complexity measures. Approximate entropy~\cite{pincus1991regularity} captures the degree of irregularity in local loss variations, with higher values indicating less predictable patterns typical of non-members. Lempel–Ziv complexity~\cite{welch1984technique} measures overall compressibility of the loss sequence, where lower compressibility (i.e., higher complexity) corresponds to non-members. Both metrics thus provide complementary views of fluctuation regularity, offering robust indicators of memorization (details in Appendix~\ref{appendix:formulation_fluctuations}).

\begin{figure}[t!]
    \centering
    \includegraphics[width=0.7\columnwidth]{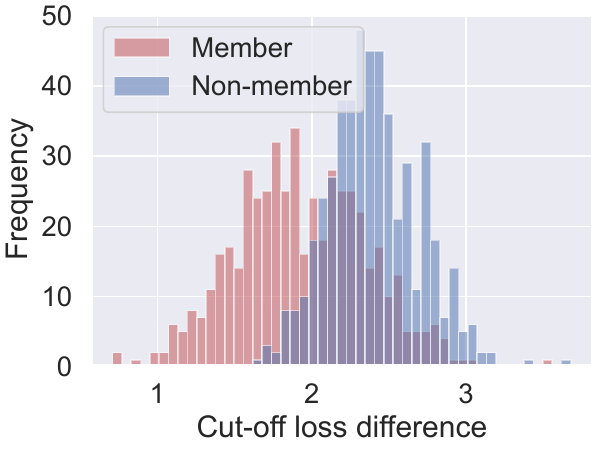}
    \caption{Distribution of loss differences after repeating inputs once ($f_{\text{Rep}}^1$) (Pythia-160M, Arxiv domain). Non-members benefit significantly more from repetition, exhibiting larger loss reductions than members, clearly highlighting memorization differences.}
    \label{fig:cut_off_loss_diff_after_repeat}
\end{figure}
\myparagraph{Amplifying signals via text repetition.}
Repeating an input provides extra context that the model can exploit during prediction. Intuitively, for unseen texts, the additional repetition supplies useful in-context cues, significantly reducing uncertainty, whereas for memorized texts, the model already ``knows'' the sequence and thus gains little benefit. For example, in Figure~\ref{fig:cut_off_loss_diff_after_repeat}, non-members exhibit much larger loss reductions after repetition compared to members. 

We capture this difference by measuring how much the loss decreases after repeating the input once or twice:
$ f_{\text{Rep}}^1(\mathbf{X}) = f(\mathbf{X};\mathcal{M}) - f(\mathbf{X};\mathcal{M}(\mathbf{X})), $
and
$ f_{\text{Rep}}^2(\mathbf{X}) = f(\mathbf{X};\mathcal{M}) - f(\mathbf{X};\mathcal{M}([\mathbf{X}, \text{`` ''}, \mathbf{X}])). $
A larger loss reduction, therefore, provides strong evidence of non-membership.

\section{MIA Test Compositions}
\label{sec:combine_signal}\label{sec:p_value_combine}

We now compose our context-aware signals (Section~\ref{sec:bag_of_signal}) into a unified membership prediction using a hypothesis-testing framework. Our composition explicitly leverages token-level, context-dependent memorization signals to yield stronger inference.

\myparagraph{Signal-level MIA tests.}
We formalize each signal as an individual hypothesis test~\cite{sankararaman2009genomic}. The null hypothesis ($H_0$) assumes that the target input $\mathbf{X}$ is a non-member, so its signal values should follow the distribution observed on held-out non-member data $D_{\text{non-mem}}$. Without loss of generality, smaller signal values indicate stronger membership evidence (signals can be negated otherwise).

For each signal $f$, we compute an empirical $p$-value following the standard Monte Carlo approach~\cite{north2002note,davison1997bootstrap,long2020pragmatic}. Specifically, we approximate the null distribution by resampling from $D_{\text{non-mem}}$ and then evaluate the extremity of the observed statistic:
\[
p_f(\mathbf{X}) = \frac{1}{|D_{\text{non-mem}}|}
\sum_{\mathbf{X}' \in D_{\text{non-mem}}}
\mathbb{1}\!\left[f(\mathbf{X}') \leq f(\mathbf{X})\right].
\]
This expression is the empirical analogue of a one-sided $p$-value: it measures the proportion of non-member samples whose statistic is at least as extreme as that of $\mathbf{X}$. A smaller $p_f(\mathbf{X})$ therefore provides stronger evidence against $H_0$ and in favor of membership. 

Our construction is a direct instantiation of the empirical $p$-value procedure widely used in permutation and bootstrap testing~\cite{north2002note,davison1997bootstrap}. Importantly, this goes beyond a mere CDF comparison: the statistic $f(\mathbf{X})$ is explicitly evaluated against the empirical null distribution under $H_0$, exactly as prescribed in standard non-parametric hypothesis testing. This framing ensures statistical validity without relying on parametric assumptions about the signal distributions, while remaining consistent with accepted practices in MIA~\cite{long2020pragmatic}.

\myparagraph{Composition of multiple tests.}
Given the individual signal-level $p$-values, we compose them into a single combined MIA test. Statistical composition methods, such as Edgington's~\cite{edgington1972additive}, Fisher's~\cite{fisher1970statistical}, or George's~\cite{mudholkar1979logit}, aggregate the evidence across multiple signals. 
For instance, Edgington’s method composes the $p$-values by summation as
${p_{\text{combined}}(\mathbf{X}) = \sum_{f \in \mathcal{F}} p_f(\mathbf{X})}$,
predicting membership if this value is below a threshold.
 Our experimental results in Section~\ref{sec:eva} validate the effectiveness of this composition strategy.

\myparagraph{Generalization with additional data.} Additional labeled member data enables stronger compositions. Appendix~\ref{sec:lr_combine} presents a learning-based approach using both member and non-member data, further improving attack performance.

\section{Experiments}
\label{sec:eva}

\begin{table*}[t!]
    \centering
    \caption{Effectiveness of attacks on the Pythia-deduped model (2.8B). We report True Positive Rate (TPR) at 1\% False Positive Rate (FPR). Higher TPR indicates better attack performance. 
    The AUC results are reported in Table~\ref{tab:mia_7_0.2_small_lr} in Appendix~\ref{appendix:additional_exp}}
    \label{tab:mia_7_0.2_small}
    \setlength{\tabcolsep}{3pt}
    \resizebox{\textwidth}{!}{%
    \large
    \begin{tabular}{lccccccc}
    \toprule
    \textbf{Attack}& \textbf{Arxiv} & \textbf{Github} & \textbf{PubMed} & \textbf{HackerNews} & \textbf{Pile-CC} & \textbf{Wikipedia} & \textbf{Mathematics} \\ 
    \midrule
    Blind~\cite{das2024blind}& 0.00 & 32.12 & 0.00 & 1.94 & 2.40 & 0.00 & \textbf{65.95} \\\hline
    LOSS~\cite{yeom2018privacy} & 14.94 & 39.84 & 18.20 & 1.06 & 4.77 & 12.37 & 12.70 \\
    Zlib~\cite{carlini2021extracting} & 10.60 & 46.12 & 14.30 & 2.05 & 5.67 & 9.44 & 8.10 \\
    Min-K\%~\cite{shi2023detecting} & 20.11 & 40.64 & 19.45 & 0.84 & 4.56 & 11.53 & 46.83 \\
    Min-K\%++~\cite{zhang2024min} & 5.20 & 31.91 & 10.44 & 1.43 & 2.87 & 10.24 & 17.94 \\
    Reference~\cite{carlini2022membership} & 5.86 & 4.68 & 1.22 & 2.63 & 5.93 & 7.36 & 0.00 \\
    Neighborhood~\cite{mattern2023membership} & 1.43 & 3.67 & 4.27 & 1.83 & 2.01 & 4.63 & 12.06 \\
    \rowcolor{gray!10} \textbf{\name (Edgington)} & 23.91 & \textbf{63.30} & 15.78 & 4.86 & \textbf{7.39} & 10.26 & 26.51 \\
    \rowcolor{gray!10} \textbf{\name (George)} & \textbf{32.00} & 61.33 & \textbf{19.94} & \textbf{5.56} & 6.76 & \textbf{13.56} & 20.63 \\
    \bottomrule
    \end{tabular}}
\end{table*}

\myparagraph{Models.} 
We evaluate MIAs against three LLM families—\emph{Pythia}~\cite{biderman2023pythia} (70M–12B parameters),\emph{Pythia-deduped} (same sizes, trained without duplicates), and \emph{GPT-Neo}~\cite{gpt-neo} (125M–2.7B parameters)—all trained on the publicly available Pile dataset~\cite{gao2020pile}.

\myparagraph{Data domains and splits.} 
We use the benchmark dataset MIMIR~\cite{duan2024membership}, which contains data from seven domains of the Pile dataset: Pile-CC (web), Wikipedia, PubMed Central, Arxiv, HackerNews, DM Mathematics, and GitHub. We evaluate using three MIMIR splits, each explicitly distinguishing members from non-members based on $n$-gram overlap.

\myparagraph{Baseline attacks.} 
We compare \name with LOSS~\cite{yeom2018privacy}, Zlib~\cite{carlini2021extracting}, Min-K\%~\cite{shi2023detecting}, Min-K\%++~\cite{zhang2024min}, Reference-based~\cite{carlini2021extracting}, and Neighborhood~\cite{mattern2023membership} (Section~\ref{sec:mia}). We use \textsf{STABLELM-BASE-ALPHA-3B-V2} as the reference model~\cite{duan2024membership}, $K=20$ for Min-K\%, and 25 neighbors for Neighborhood attack.

\myparagraph{Blind baseline.} 
To measure potential distribution shifts arising from artificial data splits, we include a blind baseline~\cite{das2024blind,meeus2024inherent}—a Naive Bayes classifier with bag-of-words features~\cite{harris1954distributional}. This classifier, trained solely on textual features (without any access to the target model) using an 80\% train and 20\% test split of the member/non-member data, and can be seen as a lower-bound baseline for MIA effectiveness in most cases.

\myparagraph{Data access for \name.} 
Our primary composition approach (Section~\ref{sec:combine_signal}) uses only non-member data for calibration. Specifically, we sample an $\alpha=30\%$ fraction of non-member test data as calibration data, with remaining non-members (70\%) and an equal number of random members forming the evaluation set.

\myparagraph{Fair comparison.} 
For fairness, baseline attacks use the same non-member calibration set as \name. Each baseline computes $p$-values from calibration data, inferring membership by thresholding these values.

\myparagraph{Signal computation in \name.} 
Detailed hyperparameter settings for each signal (Section~\ref{sec:bag_of_signal}) appear in Table~\ref{tab:individual_signals_pythia_2.8} (Appendix~\ref{appendix:additional_exp}). Token diversity is computed using the target model's tokenizer; results remain robust to common alternatives (e.g., OpenAI tokenizer, BPE~\cite{sennrich2015neural}, GPT-2~\cite{radford2019language}).

\myparagraph{Metrics.} 
We evaluate primarily using True Positive Rate (TPR) at low False Positive Rates (FPR), capturing worst-case privacy risks more accurately than average-case metrics like AUC-ROC~\cite{carlini2022membership}. AUC-ROC is also reported for completeness in Appendix~\ref{appendix:additional_exp}.

\subsection{Effectiveness of \name}
\myparagraph{Comparison with baselines.}
Table~\ref{tab:mia_7_0.2_small} compares \name with baseline attacks across seven domains using the Pythia-deduped model (2.8B). All methods are evaluated under identical conditions (i.e., datasets and calibration with non-member data only). We primarily focus on True Positive Rate (TPR) at a low False Positive Rate (1\%) and also report AUC for additional context.

\name consistently achieves higher TPR than baselines in almost all domains, clearly reflecting improved detection of memorized training points. For instance, on the Arxiv domain, \name (George) achieves a TPR of $32.00\%$, significantly surpassing the best baseline (LOSS, $14.94\%$). Similarly notable improvements occur in Github ($63.30\%$ vs. $48.61\%$) and PubMed ($19.94\%$ vs. $19.45\%$).

We note one exception. Mathematics (DM), which has limited evaluation data (only 178 points), potentially causes unreliable statistical conclusions. Additionally, as explained in Section~\ref{sec:evaluation_mia}, significant distribution shift causes even the blind (model-free) baseline to perform unusually well (TPR=$65.95\%$), clearly indicating that performance here likely reflects data distributional differences rather than model's memorization. 

In HackerNews and Pile-CC, all methods (including ours) show low performance (below $8\%$ TPR), suggesting limited memorization and inherent difficulty for MIA. We will back up our claim with experiments later (see Table~\ref{tab:effect_model_size_small}).

Other baseline methods, such as Min-K\%++~\cite{zhang2024min}, Reference-based~\cite{carlini2022membership}, and Neighborhood~\cite{mattern2023membership}, generally have stronger assumptions or higher computational cost. For example, Min-K\%++ requires full token logits, and Neighborhood performs multiple model queries per data point. Despite their higher cost, these baselines do not outperform \name in most domains. Hence, we omit Neighborhood in subsequent experiments.

Finally, if adversaries have additional access to member data, attack performance further improves, as shown in Table~\ref{tab:mia_7_0.2_small_lr} in Appendix~\ref{appendix:lr_eva}.

\myparagraph{Results on different models.} 
Table~\ref{tab:arxiv_combined_summary} summarizes attack performance specifically on the Arxiv domain, covering three widely-used model families: Pythia-deduped, Pythia, and GPT-Neo. Across all model sizes and architectures, our approach \name consistently achieves higher True Positive Rates (TPR at 1\% FPR) compared to baseline attacks. This robust performance clearly demonstrates the effectiveness of explicitly capturing token-level context-dependent memorization signals. Additional results for all data domains and detailed configurations are provided in Appendix~\ref{appendix:additional_exp}.

\myparagraph{Results on different data splits.}  
Table~\ref{tab:combined_arxiv_splits} summarizes attack effectiveness (TPR at 1\% FPR) on the Arxiv domain across varying substring-overlap splits from the MIMIR benchmark. As the allowed overlap increases from \textsf{7\_gram\_0.2} to \textsf{13\_gram\_0.8}, distinguishing members from non-members becomes substantially more challenging, and attack performance notably decreases. On the highly challenging \textsf{13\_gram\_0.8} split, most methods—including \name—perform near random guessing (1\% TPR), highlighting the difficulty of membership inference under large substring overlaps. Nonetheless, \name maintains superior or comparable performance on more distinguishable splits (\textsf{7\_gram\_0.2}, \textsf{13\_gram\_0.2}). Detailed setups and more results are in Appendix~\ref{appendix:additional_exp}.

\begin{table}[t!]
\centering
\caption{TPR at 1\% FPR on the Arxiv domain (Pythia-deduped 2.8B) across different data splits. Higher substring overlap (\textsf{13\_gram\_0.8}) makes distinguishing members more difficult.}
\label{tab:combined_arxiv_splits}
\resizebox{\columnwidth}{!}{%
\normalsize
\begin{tabularx}{1.2\columnwidth}{lYYYYYY}
\toprule
\textbf{Data Split} & \textbf{LOSS} & \textbf{Zlib} & \textbf{Min-K\%} & \textbf{Min-K\%++} & \textbf{\name}\\ 
\midrule
7\_gram\_0.2 & 14.94 & 10.60 & 20.11 & 5.20 &  \textbf{23.91} \\
13\_gram\_0.2 & 2.74 & 2.19 & 1.76 & 2.07 &  \textbf{4.00} \\
13\_gram\_0.8 & 0.50 & 0.56 & 0.43 & \textbf{1.00} &  0.41 \\
\bottomrule
\end{tabularx}}
\end{table}

\myparagraph{Efficiency of \name.}  
\name is computationally efficient, requiring only the calculation and composition of membership signals. Evaluating 1,000 samples from the Arxiv dataset using a single A100 GPU, \name completes in approximately 38 minutes. In comparison, the Neighborhood attack~\cite{mattern2023membership} takes around 500 minutes, and the Reference-based method~\cite{carlini2022membership} about 50 minutes, while simpler loss-based attacks (e.g., Zlib~\cite{carlini2021extracting}, Min-K~\cite{shi2023detecting}) take roughly 25 minutes. Thus, \name achieves superior performance at a computational cost only modestly above basic attacks, highlighting its practicality.

 \begin{table}[t!]
\centering
\caption{TPR (1\% FPR) comparison on Arxiv domain across Pythia-deduped, Pythia, and GPT-Neo models. \textbf{\name consistently outperforms all baseline attacks.}}
\label{tab:arxiv_combined_summary}
\resizebox{\columnwidth}{!}{%
\normalsize %
\begin{tabularx}{0.7\textwidth}{llYYYYYY}
\toprule
\textbf{Family} & \textbf{Size} & \textbf{LOSS} & \textbf{Zlib} & \textbf{Min-K\%} & \textbf{Min-K\%++} & \textbf{Ref} & \textbf{\name} \\ 
\midrule
\multirow{3}{*}{Pythia} 
 & 70M & 6.97 & 7.23 & 12.23 & 5.03 & 2.80 & \textbf{19.54} \\
 & 1.4B & 12.63 & 9.83 & 14.86 & 3.49 & 6.66 & \textbf{25.23} \\
 (deduped)& 2.8B & 14.94 & 10.60 & 20.11 & 5.20 & 5.86 & \textbf{23.91} \\
&6.9B & 15.14 & 13.17 & 20.37 & 4.40 & 8.29 & \textbf{28.69} \\
&12B & 15.03 & 14.86 & 21.66 & 6.31 & 8.74 & \textbf{28.06} \\
\midrule
Pythia & 2.8B & 13.14 & 10.86 & 21.54 & 5.83 & 5.14 & \textbf{24.14} \\
\midrule
\multirow{3}{*}{GPT-Neo} 
 & 125M & 9.40 & 6.94 & 9.11 & 3.91 & 2.91 & \textbf{23.09} \\
 & 1.3B & 12.74 & 11.91 & 15.09 & 4.71 & 5.86 & \textbf{25.80} \\
 & 2.7B & 16.51 & 14.40 & 21.09 & 7.91 & 7.06 & \textbf{28.57} \\
\bottomrule
\end{tabularx}}
\end{table}

\begin{table}[t!]
\centering
\caption{Impact of model size (from Pythia-deduped family) and generalization gap on \name performance. The generalization gap is the difference between training and test losses. \name's performance is TPR at 1\% FPR. Larger gaps correlate with increased memorization and thus better MIA performance.}
\label{tab:effect_model_size_small}
\resizebox{\columnwidth}{!}{%
\begin{tabular}{llccccc}
\toprule
\textbf{Domain} & \textbf{Metric} & \textbf{160M} & \textbf{1.4B} & \textbf{2.8B} & \textbf{6.9B} & \textbf{12B} \\
\midrule
\multirow{2}{*}{Arxiv} 
& Gap & 0.31 & 0.35 & 0.36 & 0.37 & 0.38 \\
& TPR & 23.37 & 25.23 & 25.89 & 28.69 & 28.06 \\
\midrule
\multirow{2}{*}{Github} 
& Gap & 1.08 & 1.02 & 1.10 & 1.01 & 1.01 \\
& TPR  & 41.81 & 54.04 & 60.21 & 55.32 & 61.38 \\
\midrule
\multirow{2}{*}{HackerNews} 
& Gap & 0.09 & 0.10 & 0.11 & 0.11 & 0.12 \\
& TPR & 2.78 & 4.99 & 4.28 & 6.45 & 6.95 \\
\midrule
\multirow{2}{*}{Pile-CC} 
& Gap & 0.07 & 0.10 & 0.10 & 0.13 & 0.15 \\
& TPR & 4.67 & 6.03 & 6.94 & 10.01 & 10.66 \\
\bottomrule
\end{tabular}}
\end{table}

\subsection{Ablation studies and insights}
\myparagraph{Impact of model size and generalization.}
Do model size and generalization ability influence \name's effectiveness? Table~\ref{tab:effect_model_size_small} compares three representative model sizes (160M, 2.8B, 12B) across various domains. We find \textit{no direct correlation} between model size and MIA effectiveness. Instead, attack performance strongly correlates with model generalization quality, quantified by the gap between train and test losses. Specifically, domains with larger generalization gaps (e.g., GitHub; gap $\approx 1.0$) reflect more significant memorization and hence, higher TPRs (up to 61.38\%). Conversely, domains with smaller gaps (e.g., HackerNews; gap $\approx 0.1$) exhibit limited memorization and lower TPRs (up to 6.95\%). Intuitively, lower generalization ability (i.e., larger gaps between train and test losses) implies increased memorization of training data, thus amplifying privacy risks. This observation further validates our previous findings that domains such as HackerNews and Pile-CC pose low privacy risks due to low memorization.

\myparagraph{Methods for combining signals.} 
Table~\ref{tab:effect_p_agg} compares various methods for combining $p$-values, including Edgington's summation~\cite{edgington1972additive}, Fisher's sum of log $p$-values~\cite{fisher1970statistical}, Pearson's negative log of complement $p$-values~\cite{pearson1933method}, and George's log-ratio method~\cite{mudholkar1979logit} (Section~\ref{sec:p_value_combine}). Performance slightly varies across domains, with no universally optimal method emerging—a finding consistent with prior statistical literature~\cite{heard2018choosing}. Edgington's simple summation approach, however, consistently achieves strong performance across domains.

\begin{table}[t!]
\centering
\caption{We show the performance of our attack when using different combination methods for combining our bag of signals. The performance is measured as TPR at 1\% FPR. We test on the Pythia-deduped 2.8b model. 
}
\label{tab:effect_p_agg}
\resizebox{\columnwidth}{!}{%
\begin{tabular}{lcccc}
\toprule
Domain &    Edgington & Fisher & Pearson & George \\\midrule
Arxiv &  25.89 &  \textbf{32.11} &   19.63 &   32.0 \\
Mathematics &  \textbf{24.92} &  20.95 &   19.37 &  20.63 \\
Github &  60.21 &  33.03 &   \textbf{67.13} &  61.33 \\
PubMed &  13.14 &  19.83 &    13.6 &  \textbf{19.94} \\
Hackernews &   4.28 &   \textbf{5.67} &    3.95 &   5.56 \\
Pile-CC&   \textbf{6.94} &   6.73 &    6.16 &   6.76 \\
\bottomrule
\end{tabular}}
\end{table}

\myparagraph{Robustness to calibration set size.} 
We evaluate \name's robustness across different calibration set sizes (parameterized by $\alpha$). Figure~\ref{fig:effect_of_attack_ratio} (Appendix~\ref{appendix:additional_exp}) shows stable performance for \name across a broad range of calibration set sizes, highlighting the method’s robustness even with limited calibration data.

\myparagraph{Individual signal effectiveness.} 
We also assess the effectiveness of individual membership signals (Section~\ref{sec:bag_of_signal}). No single signal universally performs best across all domains. For instance, token-diversity-calibrated loss ($f_{\text{Cal}}$) is particularly effective in specialized domains such as GitHub, repetition-amplified signals ($f_{\text{Rep, Cut}}^1$) excel in domains like Arxiv, and simple cut-off loss ($f_{\text{Cut}}$) performs strongly on Mathematics. This variability underscores the advantage of combining multiple signals rather than relying on any individual one. Detailed results are in Table~\ref{tab:individual_signals_pythia_2.8} in Appendix~\ref{appendix:additional_exp}.

\section{Additional related works}
Several recent studies highlight the role of prefixes in membership leakage. \citet{he2025towards} examined label-only inference, showing that contextual prefixes can reveal membership but relying on a surrogate model and mainly the first token. Similarly, \citet{meeus2025canary} used ``canary'' examples in fine-tuned models, demonstrating that high-perplexity suffixes are more easily memorized when preceded by familiar prefixes. These results support our view that ambiguous prefixes strongly drive memorization.  Other works exploit conditional likelihood differences. \citet{xie-etal-2024-recall} proposed ReCaLL, comparing log-likelihoods conditioned on non-member prefixes. Our repetition-based amplification shares this intuition but avoids external data and integrates with other signals for robustness. Since ReCaLL did not outperform the reference attack~\cite{carlini2021extracting}, we focus comparisons on that baseline.  Beyond instance-level attacks, \citet{maini2024llm} studied dataset-level inference, while \citet{puerto-etal-2025-scaling} extended to sentence-, paragraph-, dataset-, and collection-level attacks. Our work remains at the instance-level, selectively using tokens within a sample, though our signals could serve as building blocks for broader settings.  Finally, frequency-based approaches such as \citet{zhang2024pretraining} estimate token statistics from large reference corpora. This assumption is incompatible with our threat model, which operates with only limited non-member data, making direct comparison infeasible.

\section{Conclusion}
We introduce \name, a context-aware MIA framework tailored for pre-trained LLMs. Unlike traditional MIAs, \name captures token-level, context-dependent memorization overlooked by prior methods. Through a comprehensive evaluation, we demonstrate that \name significantly improves attack performance compared to existing approaches. Extending \name to evaluate fine-tuned language models and downstream applications remains a promising direction for future work.

\section{Acknowledgments}
We would like to thank all anonymous reviewers for their valuable comments.

\newpage

\section{Limitations}\label{sec:limitations}

We acknowledge several important considerations that could influence the generalization and applicability of our findings. Below, we explicitly discuss these limitations and their potential implications.

\paragraph{Evaluation language limitation.}
Our evaluations focus on pile dataset, which contains mostly English-language data, following established benchmarks. Assessing how these findings generalize to other languages remains an important direction for future research.

\paragraph{Exclusion of certain models due to benchmark limitations.}
In this paper, we did not evaluate popular large language models such as LLaMA or GPT variants because their exact training datasets are proprietary and undisclosed. Prior works typically evaluated these models using the WikiMIA benchmark~\cite{shi2023detecting}. However, recent critiques~\cite{das2024blind,meeus2024inherent} demonstrate that WikiMIA introduces substantial artificial distribution shifts between members and non-members, significantly inflating MIA performance metrics. For instance, the blind attack~\cite{das2024blind}—which relies solely on input data without querying the model—achieves nearly perfect performance (98.7\% AUC and 94.4\% TPR at 5\% FPR), far surpassing the best attacks (83.9\% AUC and 43.2\% TPR at 5\% FPR)~\cite{zhang2024min}. Due to this inherent bias, results derived from WikiMIA lack meaningful interpretability. Consequently, we chose to exclude such evaluations from our analysis. Future work should focus on developing unbiased benchmarks to rigorously evaluate membership inference attacks against models with proprietary training data.

\paragraph{Dependence on per-token loss information.}
Our membership inference attack requires access to the model's per-token loss values for a given query. While restricting access to such detailed loss information could potentially mitigate our attack, enforcing such restrictions might limit legitimate model usage scenarios and APIs that currently rely on exposing this information (e.g., for fine-tuning or debugging purposes). Understanding and balancing the trade-offs between model functionality and privacy remains an important area for future exploration.

\paragraph{Reliance on non-member calibration data.}
Our attack requires access to non-member data for calibration purposes. Although our experiments demonstrate robustness even with limited calibration data (Section~\ref{sec:eva}), performance may degrade if appropriate calibration data is scarce or significantly differs from the training data distribution. Future research should further investigate methods to reduce reliance on calibration datasets.

\section{Ethical Considerations}

We undertake this study with a strong commitment to ethical research practices and responsible disclosure. By transparently communicating our methodology, findings, and limitations, we aim to raise awareness about privacy vulnerabilities associated with large language models. Our goal is to contribute constructively to the broader community and support ongoing efforts to balance transparency, utility, and privacy, aligning our efforts with regulatory frameworks such as the EU AI Act and U.S. AI safety policies~\cite{euaiact}.

Our study exclusively utilizes publicly available datasets and models, specifically the Pythia and GPT-Neo language models (both released under the Apache 2.0 License) and the MIMIR benchmark dataset (released under the MIT License). These resources were employed strictly within their intended research purposes and license terms, and our research remains non-commercial and academic. Additionally, any artifacts created as part of this study are similarly intended solely for research purposes and are not distributed or applied beyond this scope.

We relied on comprehensive documentation provided with the Pile dataset~\cite{gao2020pile}—the training corpus for Pythia and GPT-Neo—which includes detailed disclosures on domain coverage, linguistic characteristics, and acknowledged demographic and content-related biases. MIMIR, designed as a synthetic benchmark specifically for membership inference attacks, does not contain natural language content or personally identifiable information (PII). Our research did not involve the creation of new datasets with human subjects, nor did it include any form of user data collection.

To the best of our knowledge, based on the available documentation and our intended usage, the artifacts and resources employed do not include personally identifiable information (PII) or offensive content. This aligns our work with established ethical standards concerning data privacy and content safety. For data preprocessing, modeling, and evaluation tasks, we employed widely-used, open-source software packages including Hugging Face Transformers and standard Python libraries such as NumPy and SciPy. Model-specific tokenizers and default parameter configurations were used unless explicitly stated otherwise. Lastly, AI assistants (e.g., ChatGPT) were utilized to revise this manuscript. All generated content was rigorously reviewed and finalized by the authors.

\bibliography{reference}

\appendix
\newpage
\section{Measurement of fluctuations}\label{appendix:formulation_fluctuations}

In Section~\ref{sec:bag_of_signal}, we introduced fluctuation-based signals to quantify how the uncertainty in the token-level loss sequence varies over time, capturing patterns indicative of memorization. Here, we provide formal definitions and intuitive explanations for the two fluctuation measures: \textit{Approximate Entropy} and \textit{Lempel–Ziv Complexity}.

\paragraph{Approximate entropy (ApEn).} Approximate entropy measures the unpredictability of fluctuations in the token-loss sequence by quantifying how often similar patterns recur as the sequence length increases~\cite{pincus1991regularity}. Intuitively, ApEn checks whether short segments of the sequence remain similarly close when extended slightly longer, based on a predefined similarity tolerance $r$.

Formally, given a token-loss sequence \(\{\mathcal{L}_t(x_t)\}_{t=1}^{T'}\), we define subsequences of length $m$ starting at position $t$ as:
\[
u_t^{m} = (\mathcal{L}_t(x_t), \mathcal{L}_{t+1}(x_{t+1}), \ldots, \mathcal{L}_{t+m-1}(x_{t+m-1})).
\]

We then measure the distance between two subsequences \(u_t^{m}\) and \(u_{t'}^{m}\) by the maximum absolute difference among their corresponding elements:
\begin{align}
    &d(u_t^m, u_{t'}^m) 
    \\
    &= \max_{k=1,\dots,m} \left| \mathcal{L}_{t+k-1}(x_{t+k-1}) 
       - \mathcal{L}_{t'+k-1}(x_{t'+k-1}) \right|.
\end{align}

Next, for each subsequence \(u_t^{m}\), we calculate the proportion of subsequences within a similarity threshold \(r\):
\[
C_t^{m}(r) = \frac{\#\{u_{t'}^{m} : d(u_t^{m}, u_{t'}^{m}) \leq r\}}{T'-m+1}
\]

Then, we compute the logarithmic average across all subsequences:
\[
\Phi^{m}(r) = \frac{1}{T'-m+1} \sum_{t=1}^{T'-m+1} \ln C_t^{m}(r).
\]

Finally, approximate entropy is defined as the difference between these averages at subsequence lengths \(m\) and \(m+1\):
\[
f_{\text{ApEn}}(\mathbf{X}) = \Phi^{m}(r) - \Phi^{m+1}(r).
\]

In our experiments, we choose $m=8$ and $r=0.8$, as these parameters yielded the best performance when used individually as membership inference signals.

\paragraph{Lempel–Ziv complexity (LZ complexity).} Lempel–Ziv complexity quantifies the diversity or complexity of patterns present in the token-loss sequence, inspired by compression-based methods~\cite{welch1984technique}. Intuitively, LZ complexity evaluates how many unique patterns exist in the sequence by breaking it down into the smallest number of non-repeating segments (phrases).

To apply Lempel–Ziv complexity to our continuous loss sequence, we first discretize the losses into bins, obtaining a sequence of bin indices \(\{B_1, B_2, \ldots, B_{T'}\}\), where each $B_t$ corresponds to the bin containing the token loss $\mathcal{L}_t(x_t)$. Formally, the Lempel–Ziv complexity is computed as:
\[
f_{\text{LZ}}(\mathbf{X}) = \text{LZW}(\{B_1, B_2,\ldots, B_{T'}\}),
\]

where \(\text{LZW}(\cdot)\) returns the total number of unique phrases required to describe the sequence fully according to the Lempel–Ziv–Welch compression algorithm.

These fluctuation-based signals provide robust indicators of context-dependent memorization by capturing the regularity and complexity in token-level predictions.

\section{MIA Test Composition: Learning to Compose Signals}
\label{sec:lr_combine}

In Section~\ref{sec:combine_signal}, we introduced a hypothesis-testing framework for composing multiple membership inference signals, assuming the adversary has access only to non-member data. In this appendix, we extend this approach to the more powerful setting where the adversary has access to both labeled member and non-member data, referred to collectively as the \textit{attack dataset} ($D_{\text{attack}}$). Below, we formalize this scenario as a supervised learning problem and describe our method in detail.

\paragraph{Formalization.}  
We formulate the composition of membership signals as a supervised classification task. Given a set of membership signals $\mathcal{F}=\{f_1, f_2, \ldots, f_{|\mathcal{F}|}\}$, we represent each target input $\mathbf{X}$ by a feature vector:
\[
\mathbf{X}_{\mathcal{F}} = (f_1(\mathbf{X}), f_2(\mathbf{X}), \ldots, f_{|\mathcal{F}|}(\mathbf{X})),
\]
where each element $f_i(\mathbf{X})$ is computed using the method described in Section~\ref{sec:bag_of_signal}. Our goal is to learn a model that predicts whether $\mathbf{X}$ is a member ($y=1$) or a non-member ($y=0$) of the target model’s training data.

Specifically, we use labeled member and non-member examples in $D_{\text{attack}}$ to train a binary classifier. The trained classifier then estimates the membership probability of any new query based on its computed feature vector.

\paragraph{Choice of classifier.}  
We adopt logistic regression due to its simplicity, interpretability, and efficiency. Logistic regression learns a weight vector $\mathbf{w}\in\mathbb{R}^{|\mathcal{F}|}$ that linearly combines the signal features:
\[
\hat{y} = \sigma(\langle \mathbf{X}_{\mathcal{F}}, \mathbf{w}\rangle),
\]
where $\sigma(z) = \frac{1}{1+e^{-z}}$ is the sigmoid function and $\langle\cdot,\cdot\rangle$ denotes the inner product. The model parameters $\mathbf{w}$ are trained by minimizing the standard logistic loss over the attack dataset:
\[
\min_{\mathbf{w}}-
\sum_{(\mathbf{X}, y) \in D_{\text{attack}}} \frac{y \log(\hat{y}) + (1 - y)\log(1 - \hat{y})}{|D_{\text{attack}}|},
\]
where $y$ is the true membership label of $\mathbf{X}$, and $\hat{y}$ is the predicted membership probability. After training, for a given target input $\mathbf{X}$, we first compute its feature vector and then apply the trained logistic regression model to predict membership status.

\paragraph{Dimensionality reduction with PCA.}  
For each signal defined in Section~\ref{sec:bag_of_signal}, multiple variations can exist (e.g., varying the cut-off time for the slope signal). While including all variations could potentially enhance predictive power, it may also introduce redundancy, increasing the dimensionality unnecessarily and making the classifier less stable. 

To balance predictive accuracy and complexity, we use Principal Component Analysis (PCA)~\cite{pearson1901liii} to reduce dimensionality within each group of related signals. Specifically, for each signal group, we apply PCA to compress the set of variations into a smaller number of principal components, capturing the majority of the variability within that group. These principal components then serve as inputs to the logistic regression model, providing a more compact, effective representation for membership inference.

Overall, combining supervised learning with dimensionality reduction enables our MIA framework to leverage richer available data (both members and non-members), resulting in stronger inference performance compared to the simpler hypothesis-testing approach presented in the main text (see experimental validation in Appendix~\ref{appendix:lr_eva}).

\section{Additional Experiments}\label{appendix:additional_exp}

\subsection{Extended Evaluation of \name Across Model Families}

We first provide extensive evaluations across various model families, including:

\begin{itemize}
    \item \textbf{Pythia-deduped models (70M–12B)}: Results in Table~\ref{tab:mia_7_0.2_160m}.
    \item \textbf{GPT-Neo models (125M–2.7B)}: Results in Table~\ref{tab:mia_7_0.2_gpt_neo}.
    \item \textbf{Pythia model (2.8B)}: Results in Table~\ref{tab:mia_7_0.2_pythia}.
\end{itemize}

\begin{figure*}[t!]
    \centering
    \includegraphics[width=0.8\textwidth]{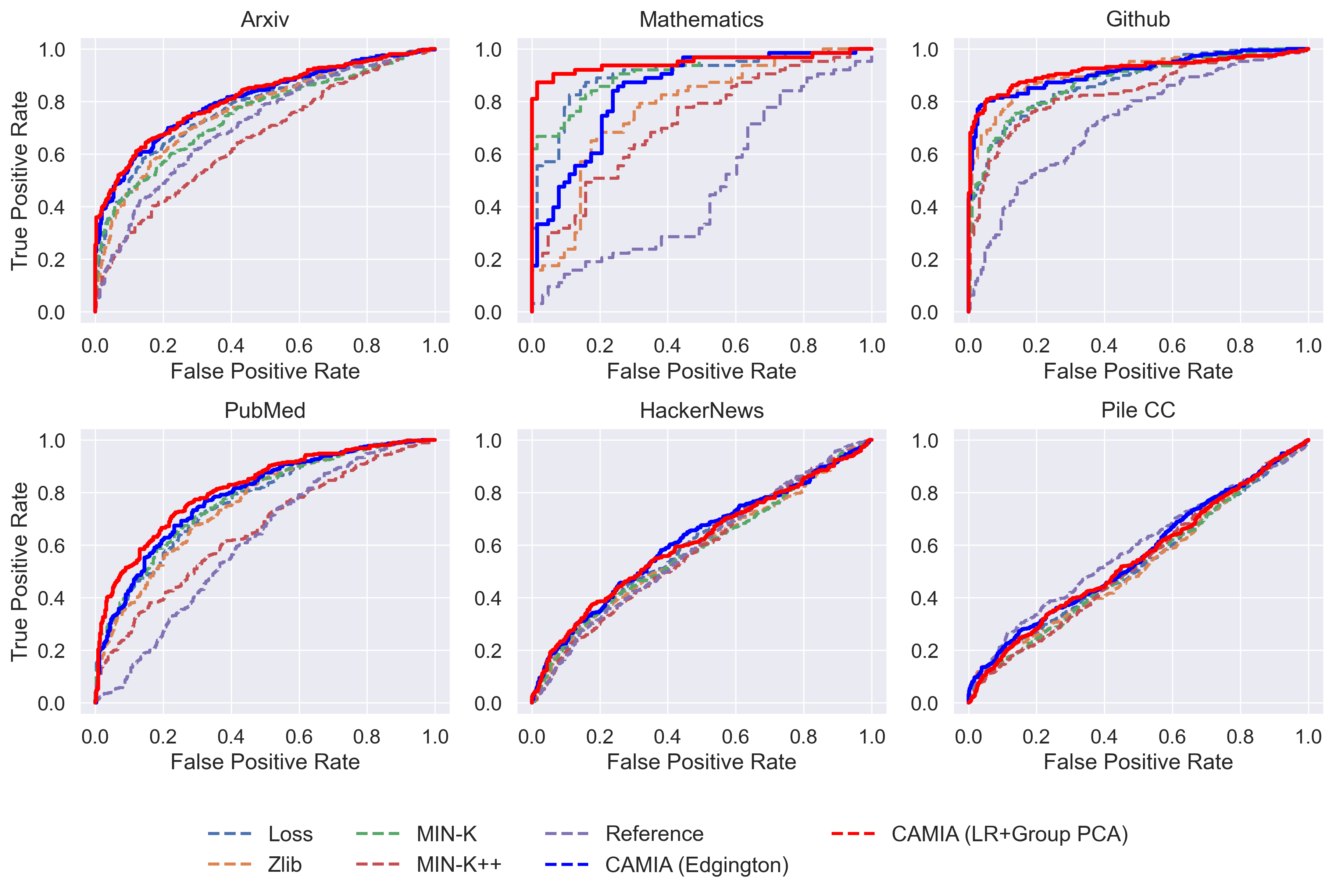}
    \caption{ROC curves comparing \name and baseline attacks (Pythia, 2.8B).}
    \label{fig:roc_2.8_7_0.2}
\end{figure*}

Consistently, our method (\name) achieves higher True Positive Rates (TPR) at low False Positive Rates (FPR), outperforming all baselines. Figure~\ref{fig:roc_2.8_7_0.2} further illustrates this superior performance via ROC curves.

\begin{table*}[t!]
\centering
\caption{Effectiveness of attacks on Pythia-deduped models with different sizes. We report the AUC and the TPR (in \%) at 1\% FPR. The results are averaged over $10$ runs across different random splits of the attack's training and test datasets. \textbf{\name consistently outperforms prior MIAs across different domains and model sizes.}}
\label{tab:mia_7_0.2_160m}
\resizebox{\textwidth}{!}{%
\small
\setlength{\tabcolsep}{2pt}
\begin{tabular}{llcccccccccccc}
\toprule
\multirow{2}{*}{Model Size} & \multirow{2}{*}{Attack/Baseline}&\multicolumn{2}{c}{Arxiv} &\multicolumn{2}{c}{Mathematics} & \multicolumn{2}{c}{Github} & \multicolumn{2}{c}{PubMed} & \multicolumn{2}{c}{HackerNews} & \multicolumn{2}{c}{Pile-CC} \\ \cmidrule(lr){3-4}\cmidrule(lr){5-6}\cmidrule(lr){7-8} \cmidrule(lr){9-10} \cmidrule(lr){11-12} \cmidrule(lr){13-14}  
& &  AUC &  TPR&  AUC &  TPR&    AUC &  TPR&  AUC &  TPR &   AUC &  TPR&  AUC &  TPR\\\midrule
&Blind~\cite{das2024blind}&0.76 & 0.0&0.93 & 65.95&0.84 & 32.12&0.75 & 0.0&0.53 & 1.94&0.55 & 2.4\\\hline
\multirow{8}{*}{70M}&LOSS~\cite{yeom2018privacy}  &                   0.73 &                   6.97 &                            0.95 &                           78.73 &                    0.82 &                   19.52 &                            0.79 &                           15.47 &                        0.58 &                        3.05 &                     0.53 &                     1.94 \\
&Zlib~\cite{carlini2021extracting} &                   0.73 &                   7.23 &                            0.82 &                           28.41 &                    0.86 &                   48.94 &                            0.78 &                           14.01 &                        0.57 &                        2.72 &                     0.51 &                     3.23 \\
& Min-K\%~\cite{shi2023detecting} &                   0.68 &                  12.23 &                            0.94 &                           75.56 &                    0.81 &                   19.31 &                            0.77 &                           14.04 &                        0.56 &                        3.55 &                     0.53 &                     1.89 \\
&Min-K\%++~\cite{zhang2024min}  &                   0.56 &                   5.03 &                            0.74 &                            6.83 &                    0.72 &                    7.66 &                            0.63 &                            4.65 &                        0.55 &                        1.24 &                     0.52 &                     1.74 \\
&Reference~\cite{carlini2021extracting}  &                   0.52 &                   2.80 &                            0.63 &                            2.06 &                    0.68 &                    4.31 &                            0.66 &                            6.16 &                        0.52 &                        0.46 &                     0.50 &                     2.13 \\\cmidrule(lr){2-14}
& \name (Edgington)  &                   0.77 &                  19.54 &                            0.93 &                           69.68 &                    0.85 &                   43.03 &                            0.81 &                           16.57 &                        \textbf{0.59 }&                        \textbf{4.83} &                     \textbf{0.53} &                    \textbf{ 4.01} \\

& \name (LR+ Group PCA) &                   \textbf{0.79 }&                  \textbf{23.37 }&                            \textbf{0.95} &                          \textbf{ 80.95} &                    \textbf{0.87} &                   \textbf{53.46} &                           \textbf{ 0.83 }&                          \textbf{ 27.85 }&                       \textbf{ 0.59} &                        3.64 &                     0.51 &                     1.09 \\

\hline\hline
\multirow{8}{*}{160M}& LOSS~\cite{yeom2018privacy} &                   0.74 &                   7.26 &                            0.94 &                           70.32 &                    0.83 &                   24.31 &                            0.79 &                           18.28 &                        0.57 &                        2.91 &                     \textbf{0.54} &                     2.70 \\
&Zlib~\cite{carlini2021extracting}  &                   0.74 &                   6.06 &                            0.81 &                           21.59 &                    0.87 &                   45.48 &                            0.78 &                           16.74 &                        0.57 &                        2.76 &                     0.52 &                     3.37 \\
& Min-K\%~\cite{shi2023detecting} &                   0.69 &                   8.49 &                            0.92 &                           68.89 &                    0.82 &                   25.74 &                            0.78 &                           17.67 &                        0.55 &                        2.67 &                     0.53 &                     2.99 \\
&Min-K\%++~\cite{zhang2024min}  &                   0.53 &                   2.14 &                            0.76 &                           16.51 &                    0.72 &                   10.48 &                            0.62 &                            8.02 &                        0.53 &                        1.10 &                     0.52 &                     2.30 \\
&Reference~\cite{carlini2021extracting} &                   0.57 &                   1.09 &                            0.62 &                            0.16 &                    0.68 &                    3.51 &                            0.68 &                            4.04 &                        0.51 &                        1.04 &                     0.52 &                     2.64 \\\cmidrule(lr){2-14}
&\name (Edgington) &                   0.79 &                  23.37 &                            0.90 &                           31.11 &                    0.87 &                   41.81 &                            0.81 &                           21.25 &                        0.59 &                        2.78 &                     \textbf{0.54 }&                     \textbf{4.67} \\
&\name (LR+Group PCA) &                  \textbf{0.80 }&                  \textbf{24.74 }&                           \textbf{ 0.95} &                           \textbf{73.97} &                    \textbf{0.88} &                   \textbf{56.91} &                            \textbf{0.83 }&                          \textbf{ 30.93 }&                        \textbf{0.59 }&                        \textbf{4.26 }&                     0.53 &                     1.40 \\
\hline\hline
\multirow{8}{*}{1.4B}& LOSS~\cite{yeom2018privacy} &                   0.77 &                  12.63 &                            0.92 &                           43.49 &                    0.86 &                   30.05 &                            0.78 &                           16.16 &                        0.59 &                        1.99 &                     0.55 &                     4.56 \\
& Zlib~\cite{carlini2021extracting} &                   0.77 &                   9.83 &                            0.80 &                           15.24 &                    0.89 &                   36.38 &                            0.77 &                           13.95 &                        0.58 &                        2.19 &                     0.54 &                     5.91 \\
& Min-K\%~\cite{shi2023detecting} &                   0.74 &                  14.86 &                            0.93 &                           67.14 &                    0.85 &                   29.95 &                            0.78 &                           18.05 &                        0.57 &                        2.14 &                     0.55 &                     4.70 \\
& Min-K\%++~\cite{zhang2024min} &                   0.64 &                   3.49 &                            0.75 &                           15.87 &                    0.81 &                   22.55 &                            0.63 &                            8.08 &                        0.55 &                        1.52 &                     0.55 &                     3.96 \\
&Reference~\cite{carlini2021extracting} &                   0.71 &                   6.66 &                            0.50 &                            1.27 &                    0.72 &                    0.96 &                            0.67 &                            1.54 &                        0.54 &                        1.39 &                     \textbf{0.59} &                     5.90 \\\cmidrule(lr){2-14}
&      \name (Edgington) &                   \textbf{0.81} &                  25.23 &                            0.83 &                           11.90 &                    0.89 &                   54.04 &                            0.79 &                           14.22 &                        0.60 &                        \textbf{4.99 }&                     0.55 &                    \textbf{ 6.03} \\
& \name (LR+ Group PCA) &                  \textbf{ 0.81} &                  \textbf{31.23} &                            \textbf{0.95} &                           \textbf{71.90 }&                    \textbf{0.91 }&                   \textbf{57.77} &                            \textbf{0.82} &                           \textbf{26.22} &                        \textbf{0.60} &                        4.55 &                     0.55 &                     2.74 \\

\hline\hline
\multirow{8}{*}{2.8B}&LOSS~\cite{yeom2018privacy} &                   0.78 &                  14.11 &                            0.91 &                           19.21 &                    0.87 &                   39.68 &                            0.78 &                           18.28 &                        \textit{0.60 }&                        2.03 &                     0.55 &                     4.63 \\
&Zlib~\cite{carlini2021extracting} &                   0.77 &                  10.86 &                            0.80 &                           11.43 &                    0.90 &                   42.02 &                            0.77 &                           14.51 &                        0.59 &                        2.49 &                     0.54 &                     5.83 \\
&Min-K\%~\cite{shi2023detecting} &                   0.75 &                  20.63 &                            0.92 &                           \textit{54.60} &                    0.87 &                   40.27 &                            0.78 &                           20.09 &                        0.58 &                        1.24 &                     0.55 &                     4.71 \\
&Min-K\%++~\cite{zhang2024min} &                   0.65 &                   5.71 &                            0.72 &                           19.84 &                    0.84 &                   31.49 &                            0.66 &                            9.80 &                        0.57 &                        1.52 &                     0.54 &                     3.27 \\
&Reference~\cite{carlini2021extracting} &                   0.71 &                   6.46 &                            0.45 &                            0.79 &                    0.72 &                    4.79 &                            0.63 &                            1.60 &                        0.57 &                        3.00 &                     \textbf{0.59} &                     6.34 \\\cmidrule(lr){2-14}
&\name (Edgington) &                   \textbf{0.81} &                  25.89 &                            0.83 &                           24.92 &                    0.90 &                   60.21 &                            0.79 &                           13.14 &                        \textbf{0.61} &                        4.28 &                     0.55 &                     \textbf{6.94} \\
&\name (LR + Group PCA) &                   \textbf{0.81} &                 \textbf{32.89} &                            \textbf{0.95 }&                          \textbf{ 72.22} &                   \textbf{0.91} &                   \textbf{64.57 }&                           \textbf{ 0.82} &                           \textbf{26.72} &                        0.60&                        \textbf{4.46} &                     0.54 &                     2.70 \\\hline\hline

\multirow{8}{*}{6.9B}& LOSS~\cite{yeom2018privacy} &                   0.78 &                  15.14 &                            0.92 &                           26.35 &                    0.87 &                   33.88 &                            0.78 &                           16.51 &                        0.60 &                        1.85 &                     0.57 &                     6.61 \\
& Zlib~\cite{carlini2021extracting} &                   0.78 &                  13.17 &                            0.80 &                           12.38 &                    0.90 &                   38.46 &                            0.77 &                           13.26 &                        0.59 &                        2.78 &                     0.55 &                     7.54 \\
& Min-K\%~\cite{shi2023detecting} &                   0.75 &                  20.37 &                            0.92 &                           60.79 &                    0.87 &                   34.95 &                            0.78 &                           18.98 &                        0.59 &                        2.10 &                     0.57 &                     6.20 \\
& Min-K\%++~\cite{zhang2024min} &                   0.65 &                   4.40 &                            0.73 &                           17.78 &                    0.84 &                   25.48 &                            0.67 &                            8.55 &                        0.58 &                        1.92 &                     0.56 &                     5.13 \\
&Reference~\cite{carlini2021extracting}&                   0.72 &                   8.29 &                            0.46 &                            2.06 &                    0.64 &                    0.64&                            0.60 &                            1.31 &                        0.58 &                        1.77 &                     \textbf{0.64 } &                     9.87 \\\cmidrule(lr){2-14}
&\name (Edgington)&                   \textbf{0.82 }&                  28.69 &                            0.86 &                           29.05 &                    0.90 &                   55.32 &                            0.79 &                           11.89 &                       \textbf{ 0.61} &                       \textbf{ 6.45} &                     0.58 &                    \textbf{10.01} \\

&\name (LR + Group PCA) &                   \textbf{0.82} &                 \textbf{ 33.23} &                            \textbf{0.95} &                           \textbf{70.79} &                    \textbf{0.91} &                   \textbf{63.72} &                           \textbf{ 0.82} &                          \textbf{ 24.53} &                        \textbf{0.61} &                        5.12 &                     0.57 &                     4.31 \\\hline\hline
\multirow{8}{*}{12B}&
LOSS~\cite{yeom2018privacy} &                   0.79 &                  15.03 &                            0.92 &                           17.30 &                    0.88 &                   35.05 &                            0.77 &                           16.54 &                        0.61 &                        2.14 &                     0.58 &                     7.14 \\
& Zlib~\cite{carlini2021extracting} &                   0.78 &                  14.86 &                            0.81 &                            9.37 &                    0.91 &                   36.70 &                            0.77 &                           11.77 &                        0.60 &                        3.07 &                     0.56 &                     8.57 \\
& Min-K\%~\cite{shi2023detecting} &                   0.77 &                  21.66 &                            0.92 &                           51.11 &                    0.88 &                   35.21 &                            0.78 &                           20.99 &                        0.60 &                        2.43 &                     0.58 &                     6.49 \\
& Min-K\%++~\cite{zhang2024min} &                  0.68 &                   6.31 &                            0.70 &                           22.70 &                    0.86 &                   27.23 &                            0.67 &                            9.80 &                        0.59 &                        1.96 &                     0.58 &                     6.51 \\
&Reference~\cite{carlini2021extracting}&                    0.73 &                   8.74 &                            0.45 &                            0.48 &                    0.61 &                    0.69 &                            0.58 &                            1.05 &                        0.61 &                        2.72 &                     0.67 &                    10.57 \\\cmidrule(lr){2-14}
&\name (Edgington)             &                   \textbf{0.82} &                  28.06 &                            0.85 &                           27.62 &                    0.91 &                   61.38 &                            0.79 &                           11.77 &                        \textbf{0.61} &                        \textbf{6.95} &                    \textbf{ 0.59 }&                    \textbf{10.66 }\\
&\name (LR + Group PCA) &                   \textbf{0.82 }&                 \textbf{ 36.06} &                            \textbf{0.95} &                           \textbf{69.84 }&                   \textbf{ 0.92} &                   \textbf{63.78} &                            \textbf{0.82} &                           \textbf{21.28} &                        \textbf{0.61 }&                        5.74 &                     0.58 &                     4.89 \\
\bottomrule
\end{tabular}}
\end{table*}

\begin{table*}[t!]
\centering
\caption{Effectiveness of attacks on GPT-Neo models with different sizes. We report the AUC and the TPR (in \%) at 1\% FPR. The results are averaged over $10$ runs across different random splits of the attack's training and test datasets. \textbf{\name consistently outperforms prior MIAs across different domains and model sizes.}}
\label{tab:mia_7_0.2_gpt_neo}
\resizebox{\textwidth}{!}{%
\small
\setlength{\tabcolsep}{2pt}
\begin{tabular}{llcccccccccccc}
\toprule
\multirow{2}{*}{Model Size} & \multirow{2}{*}{Attack}&\multicolumn{2}{c}{Arxiv} &\multicolumn{2}{c}{Mathematics} & \multicolumn{2}{c}{Github} & \multicolumn{2}{c}{PubMed} & \multicolumn{2}{c}{HackerNews} & \multicolumn{2}{c}{Pile-CC} \\ \cmidrule(lr){3-4}\cmidrule(lr){5-6}\cmidrule(lr){7-8} \cmidrule(lr){9-10} \cmidrule(lr){11-12} \cmidrule(lr){13-14}  
& &  AUC &  TPR&  AUC &  TPR&    AUC &  TPR&  AUC &  TPR &   AUC &  TPR&  AUC &  TPR\\\midrule
\multirow{8}{*}{125M}& LOSS~\cite{yeom2018privacy} &                       0.76 &                   9.40 &                           \textbf{ 0.95 }&                           \textbf{77.94} &                    0.83 &                   24.84 &                            0.81 &                           19.53 &                        0.57 &                        1.59 &                     0.53 &                     2.29 \\
& Zlib~\cite{carlini2021extracting}  &                   0.76 &                   6.94 &                            0.82 &                           27.62 &                    0.86 &                   39.68 &                            0.79 &                           21.48 &                        0.57 &                        2.25 &                     0.51 &                     2.53 \\
& Min-K\%~\cite{shi2023detecting}  &                   0.72 &                   9.11 &                            0.94 &                           75.24 &                    0.82 &                   25.43 &                            0.80 &                           21.13 &                        0.56 &                        1.52 &                     0.53 &                     2.61 \\
& Min-K\%++~\cite{zhang2024min} &                   0.62 &                   3.91 &                            0.70 &                           16.35 &                    0.78 &                   18.78 &                            0.67 &                           10.29 &                        0.54 &                        2.14 &                     0.52 &                     2.13 \\
&Reference~\cite{carlini2021extracting}  &                   0.65 &                   2.91 &                            0.56 &                            0.32 &                    0.67 &                    1.12 &                            0.73 &                            6.40 &                        0.51 &                        0.84 &                     0.51 &                     2.29 \\\cmidrule(lr){2-14}
&\name (Edgington)&                   0.81 &                  23.09 &                            0.88 &                           16.51 &                    0.87 &                   50.53 &                            0.82 &                           19.94 &                        \textbf{0.59 }&                        2.41 &                     \textbf{0.54} &                    \textbf{ 4.10} \\
&\name (LR+Group PCA) &                   \textbf{0.82} &                  \textbf{28.00 }&                            \textbf{0.95 }&                           77.30 &                    \textbf{0.89 }&                   \textbf{65.85} &                            \textbf{0.84} &                           \textbf{28.90} &                        0.57 &                        \textbf{3.47} &                     0.52 &                     1.50 \\\hline\hline
\multirow{8}{*}{1.3B}& LOSS~\cite{yeom2018privacy} &                   0.78 &                  12.74 &                            0.93 &                           63.02 &                    0.86 &                   39.10 &                            0.80 &                           18.81 &                        0.59 &                        1.74 &                     0.54 &                     4.56 \\
& Zlib~\cite{carlini2021extracting} &                   0.78 &                  11.91 &                            0.80 &                           14.76 &                    0.88 &                   51.28 &                            0.78 &                           19.48 &                        0.58 &                        2.19 &                     0.53 &                     4.24 \\
& Min-K\%~\cite{shi2023detecting}  &                   0.75 &                  15.09 &                            0.93 &                           70.48 &                    0.86 &                   38.94 &                            0.80 &                           22.24 &                        0.57 &                        1.96 &                     0.54 &                     4.10 \\
& Min-K\%++~\cite{zhang2024min} &                   0.66 &                   4.71 &                            0.71 &                           26.03 &                    0.81 &                   33.35 &                            0.68 &                            9.01 &                        0.56 &                        1.88 &                     0.53 &                     2.99 \\
&Reference~\cite{carlini2021extracting}  &                   0.71 &                   5.86 &                            0.49 &                            1.43 &                    0.66 &                    1.97 &                            0.70 &                            1.31 &                        0.52 &                        1.66 &                    \textbf{ 0.55 }&                     4.60 \\\cmidrule(lr){2-14}
&\name (Edgington) &                   \textbf{0.82} &                  25.80 &                            0.83 &                           18.73 &                    0.90 &                   62.50 &                            0.81 &                           17.70 &                        \textbf{0.60} &                        \textbf{4.17 }&                     \textbf{0.55} &                     \textbf{6.16 }\\
&\name (LR+Group PCA)&                   \textbf{0.82} &                  \textbf{32.23 }&                            \textbf{0.95} &                           \textbf{74.44 }&                   \textbf{ 0.91} &                   \textbf{65.96} &                           \textbf{ 0.83} &                           \textbf{27.94} &                        0.59 &                        3.75 &                     0.54 &                     2.80 \\\hline\hline
\multirow{8}{*}{2.7B}&LOSS~\cite{yeom2018privacy}  &                   0.79 &                  16.51 &                            0.93 &                           55.71 &                    0.87 &                   41.86 &                            0.80 &                           21.25 &                        \textbf{0.59 }&                        1.61 &                     0.55 &                     4.97 \\
& Zlib~\cite{carlini2021extracting} &                   0.78 &                  14.40 &                            0.81 &                           15.87 &                    0.89 &                   50.32 &                            0.78 &                           18.63 &                        0.58 &                        2.43 &                     0.54 &                     5.34 \\
& Min-K\%~\cite{shi2023detecting} &                   0.76 &                  21.09 &                            0.93 &                           69.68 &                    0.87 &                   42.18 &                            0.80 &                           23.46 &                        0.57 &                        1.79 &                     0.55 &                     4.64 \\
& Min-K\%++~\cite{zhang2024min}&                   0.66 &                   7.91 &                            0.72 &                           27.14 &                    0.83 &                   34.20 &                            0.69 &                           12.18 &                        0.57 &                        1.96 &                     0.54 &                     3.79 \\
&Reference~\cite{carlini2021extracting}  &                   0.72 &                   7.06 &                            0.52 &                            0.32 &                    0.65 &                    1.54 &                            0.69 &                            1.66 &                        0.52 &                        1.88 &                    \textbf{ 0.57 }&                     5.60 \\\cmidrule(lr){2-14}
&\name (Edgington)  &                   0.82 &                  28.57 &                            0.86 &                           21.75 &                    0.91 &                   60.59 &                            0.81 &                           15.84 &                        \textbf{0.59 }&                        2.69 &                     0.56 &                     \textbf{5.97 }\\
&\name (LR+Group PCA) &                   \textbf{0.83 }&                  \textbf{37.03 }&                            \textbf{0.95} &                           \textbf{73.65 }&                    \textbf{0.92 }&                   \textbf{67.50 }&                            \textbf{0.83 }&                           \textbf{23.63 }&                        0.58 &                        \textbf{4.13} &                     0.55 &                     3.23 \\
\bottomrule
\end{tabular}}
\end{table*}

\begin{table*}[t!]
\centering
\caption{Effectiveness of attacks on Pythia models with 2.8B. We report the AUC and the TPR (in \%) at 1\% FPR. The results are averaged over $10$ runs across different random splits of the attack's training and test datasets.}
\label{tab:mia_7_0.2_pythia}
\resizebox{\textwidth}{!}{%
\setlength{\tabcolsep}{2pt}
\begin{tabular}{llcccccccccccc}
\toprule
\multirow{2}{*}{} & \multirow{2}{*}{Attack}&\multicolumn{2}{c}{Arxiv} &\multicolumn{2}{c}{Mathematics} & \multicolumn{2}{c}{Github} & \multicolumn{2}{c}{PubMed} & \multicolumn{2}{c}{HackerNews} & \multicolumn{2}{c}{Pile-CC} \\ \cmidrule(lr){3-4}\cmidrule(lr){5-6}\cmidrule(lr){7-8} \cmidrule(lr){9-10} \cmidrule(lr){11-12} \cmidrule(lr){13-14}  
& &  AUC &  TPR&  AUC &  TPR&    AUC &  TPR&  AUC &  TPR &   AUC &  TPR&  AUC &  TPR\\\midrule
&Loss~\cite{yeom2018privacy}  &                   0.78 &                  13.14 &                            0.91 &                           20.63 &                    0.87 &                   37.02 &                            0.77 &                           18.95 &                        0.60 &                        1.28 &                     0.54 &                     4.86 \\
& Zlib~\cite{carlini2021extracting}  &                   0.77 &                  10.86 &                            0.79 &                           11.11 &                    0.90 &                   42.13 &                            0.76 &                           15.00 &                        0.58 &                        2.08 &                     0.53 &                     6.49 \\
& MIN-K~\cite{shi2023detecting} &                   0.75 &                  21.54 &                            0.92 &                           53.65 &                    0.87 &                   37.66 &                            0.78 &                           20.47 &                        0.58 &                        1.06 &                     0.54 &                     5.34 \\
& MIN-K++~\cite{zhang2024min}  &                   0.65 &                   5.83 &                            0.70 &                           17.46 &                    0.84 &                   30.85 &                            0.66 &                            9.71 &                        0.57 &                        1.46 &                     0.54 &                     3.29 \\
&Reference~\cite{carlini2021extracting} &                   0.71 &                   5.14 &                            0.44 &                            1.27 &                    0.73 &                    4.89 &                            0.62 &                            1.28 &                        0.57 &                        2.65 &                     \textbf{0.58} &                     \textbf{6.60}\\\cmidrule(lr){2-14}
&\name (Edgington)  &                  \textbf{ 0.81} &                  24.14 &                            0.82 &                           27.62 &                   \textbf{ 0.91} &                   59.63 &                            0.79 &                           11.95 &                        \textbf{0.61} &                        3.16 &                     0.55 &                     6.49 \\
&\name (LR + Group PCA)&                   \textbf{0.81} &                  \textbf{33.49} &                            \textbf{0.95} &                           \textbf{72.70} &                    \textbf{0.91 }&                   \textbf{65.74} &                            \textbf{0.82} &                           \textbf{26.66} &                        0.60 &                        \textbf{4.17 }&                     0.54 &                     2.81 \\
\bottomrule
\end{tabular}}
\end{table*}

\subsection{Robustness to Calibration Dataset Size}

\begin{figure*}[t!]
    \centering
    \includegraphics[width=0.6\textwidth]{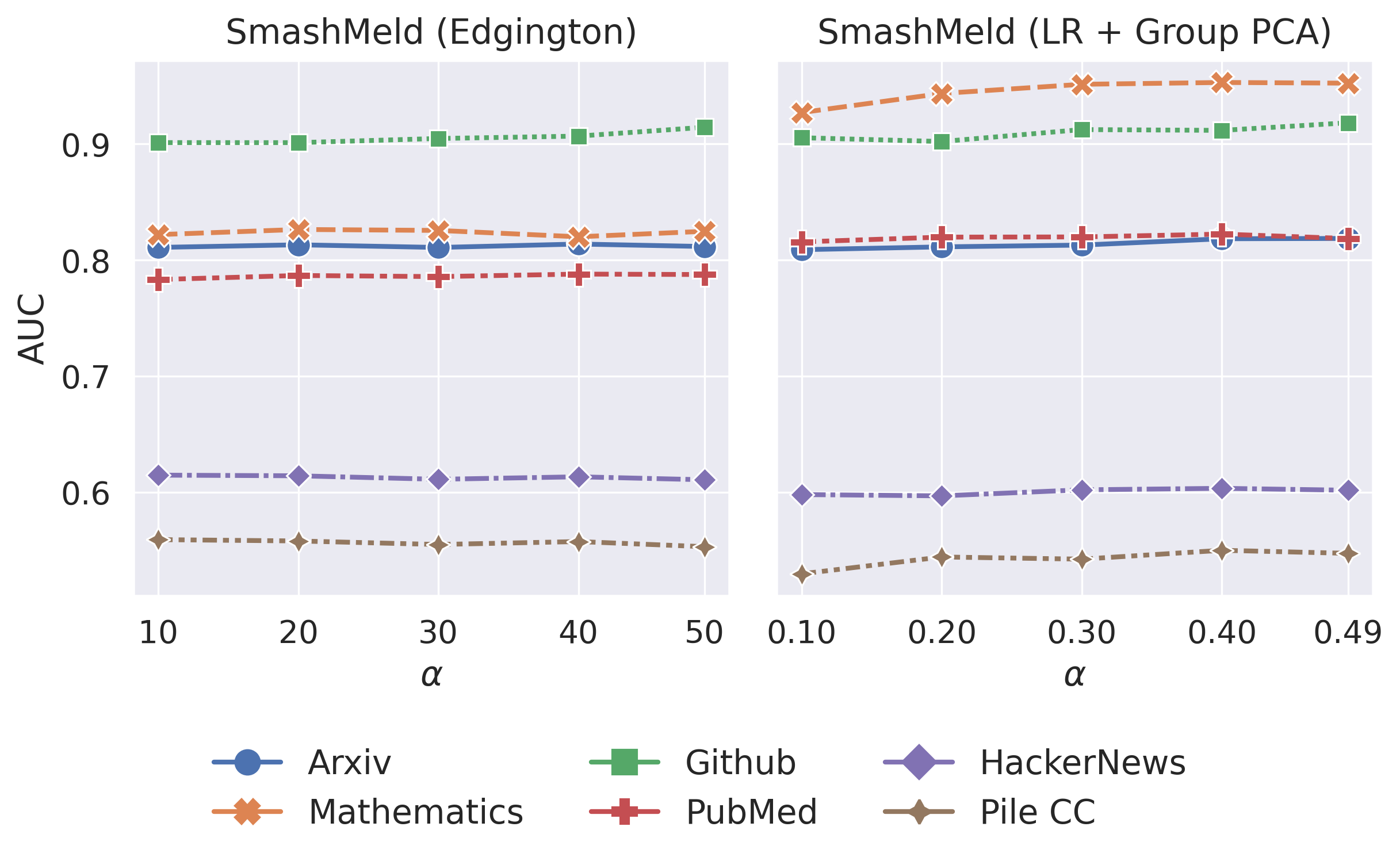}
    \caption{Effect of calibration set size ($\alpha$) on \name’s performance (Pythia-deduped, 2.8B).}
    \label{fig:effect_of_attack_ratio}
\end{figure*}

Recall from Section~\ref{sec:p_value_combine} that \name primarily uses non-member data for calibration. Figure~\ref{fig:effect_of_attack_ratio} demonstrates stable attack performance (AUC) across varying calibration sizes ($\alpha$). Additionally, when member data is also accessible, logistic regression (LR)-based signal combination (introduced in Appendix~\ref{sec:lr_combine}) further improves performance, highlighting the benefit of richer training data.

\subsection{Effectiveness Across Splits with Different Overlap Levels}

Tables~\ref{tab:mia_13_0.2_2.8b} and~\ref{tab:mia_13_0.8_2.8b} provide results on MIMIR splits (\textsf{13\_gram\_0.2}, \textsf{13\_gram\_0.8}) with varying membership overlap. Increased overlap reduces overall effectiveness due to ambiguity in membership. Nonetheless, \name consistently maintains superior performance in clearly separable cases (e.g., \textsf{13\_gram\_0.2}).

\begin{table*}[t!]
\centering
\caption{Effectiveness of attacks on Pythia models with size 2.8B on multiple domains for \textsf{ngram\_13\_0.2} split. We report the AUC and the TPR at 1\% FPR.}
\label{tab:mia_13_0.2_2.8b}
\resizebox{\textwidth}{!}{%
\begin{tabular}{lcccccccccccc}
\toprule
& \multicolumn{2}{c}{Arxiv} &\multicolumn{2}{c}{Mathematics} & \multicolumn{2}{c}{Github} & \multicolumn{2}{c}{Hackernews} & \multicolumn{2}{c}{Pile CC} & \multicolumn{2}{c}{Pubmed Central}\\
Method &  AUC &  TPR&  AUC &  TPR&    AUC &  TPR&  AUC &  TPR &   AUC &  TPR&  AUC &  TPR\\\midrule
Blind & 0.53 & 0.41&0.68 & 0.0&0.8 & 16.08&0.52 & 1.83&0.51 & 0.54&0.53 & 1.44\\\hline
Loss &                    \textbf{0.57 }&                    2.74 &                             \textbf{0.68} &                             1.08 &                     0.81 &                    37.20 &                             0.53 &                             1.97 &                         0.52 &                         0.79 &                      0.51 &                      2.74 \\
Zlib &                    0.56 &                    2.19 &                             0.65 &                             2.21 &                     0.84 &                    48.92 &                             0.53 &                             1.07 &                         0.52 &                         1.32 &                      0.52 &                      2.64 \\
MIN-K &                    0.56 &                    1.76 &                             0.65 &                             5.66 &                     0.81 &                    35.85 &                             0.53 &                             1.30 &                         0.53 &                         0.69 &                      0.52 &                      1.97 \\
MIN-K++ &                    0.56 &                    2.07 &                             0.59 &                             4.15 &                     0.79 &                    31.06 &                             0.52 &                             1.71 &                         0.53 &                         1.11 &                      0.52 &                      2.77 \\
\hline
\name (Edgington) &                    \textbf{0.57 }&                    \textbf{4.00} &                             0.64 &                             \textbf{5.88} &                     \textbf{0.85} &                    54.58 &                             0.53 &                             1.71 &                         0.53 &                         1.11 &                      0.52 &                      \textbf{2.81} \\
\name (LR + Group PCA) &                    0.56 &                    2.81 &                             \textbf{0.68} &                             2.10 &                     \textbf{0.85} &                    \textbf{59.61} &                             0.53 &                             1.91 &                         0.50 &                         1.18 &                     \textbf{ 0.53} &                      2.50 \\
\bottomrule
\end{tabular}}
\end{table*}

\begin{table*}[h]
\centering
\caption{Effectiveness of attacks on Pythia-2.B models on multiple domains for \textsf{ngram\_13\_0.8} split. We report the AUC and the TPR at 1\% FPR. }
\label{tab:mia_13_0.8_2.8b}
\setlength{\tabcolsep}{2pt}
\resizebox{0.9\textwidth}{!}{%
\begin{tabular}{lcccccccccc}
\toprule
& \multicolumn{2}{c}{Arxiv} &\multicolumn{2}{c}{Mathematics} & \multicolumn{2}{c}{Hackernews} & \multicolumn{2}{c}{Pile CC} & \multicolumn{2}{c}{Pubmed Central}\\
Method &  AUC &  TPR&  AUC &  TPR&    AUC &  TPR&  AUC &  TPR&  AUC &  TPR\\\midrule
Bline & 0.48 & 0.5&0.51 & 0.2&0.51 & 0.35&0.54 & 1.03&0.51 & 0.46\\\hline
Loss &                    0.52 &                    0.50 &                             0.48 &                             1.14 &                             0.49 &                             0.77 &                         0.51 &                         0.67 &                       0.5 &                      0.83 \\
   Zlib &                    0.51 &                    0.56 &                             0.48 &                             1.06 &                             0.50 &                             0.66 &                         0.51 &                         1.20 &                       0.5 &                      1.09 \\
  MIN-K &                    0.52 &                    0.43 &                             0.49 &                             0.44 &                             0.50 &                             0.74 &                         0.52 &                         0.73 &                       0.5 &                      0.83 \\
MIN-K++ &                    0.53 &                    1.00 &                             0.50 &                             1.24 &                             0.51 &                             1.16 &                         0.53 &                         1.17 &                       0.5 &                      1.20 \\\hline

\name (Edgington) &                    0.52 &                    0.41 &                             0.52 &                             1.64 &                             0.53 &                             1.20 &                         0.51 &                         1.09 &                      0.51 &                      1.86 \\
\name (LR + Group PCA) &                    0.50 &                    1.29 &                             0.50 &                             1.06 &                             0.50 &                             1.07 &                         0.49 &                         1.01 &                      0.48 &                      0.93 \\
\bottomrule
\end{tabular}
}
\end{table*}

\subsection{Individual Signal Performance Analysis}

We present detailed results on individual membership signals from Section~\ref{sec:bag_of_signal} in Tables~\ref{tab:individual_signals_pythia_2.8} and~\ref{tab:individual_signals_pythia_2.8_2}. Specifically, we consider:

\begin{itemize}
    \item \textbf{Cut-off loss ($f_{\text{Cut}}$)}: Evaluated with $T'=200, 300, T$; repetition-amplified versions $f_{\text{Rep,Cut}}^1, f_{\text{Rep,Cut}}^2$.
    \item \textbf{Token diversity calibrated loss ($f_{\text{Cal}}$)}.
    \item \textbf{Perplexity ($f_{\text{PPL}}, f_{\text{Cal,PPL}}$)}: Standard and calibrated perplexity signals.
    \item \textbf{Robust counting signals ($f_{\text{CB}}, f_{\text{CBM}}, f_{\text{CBPM}}$)}.
    \item \textbf{Lempel–Ziv complexity ($f_{\text{LZ}}$)}.
    \item \textbf{Slope of loss sequence ($f_{\text{Slope}}$)}.
    \item \textbf{Approximate entropy ($f_{\text{ApEn}}$)}.
\end{itemize}

No individual signal universally excels, emphasizing the value of combining multiple signals.

\begin{table*}[t!]
\centering
\caption{Performance of using each individual signal. We show the TPR at 1\% FPR for the Pythia-deduped model (2.8B). }
\label{tab:individual_signals_pythia_2.8}
\resizebox{0.9\textwidth}{!}{%
\begin{tabular}{llrrrrrr}
\toprule
Feature& Configurations&  Arxiv &  Mathematics&  GitHub &  PubMed &  HackerNews &  Pile-CC\\
\midrule
\multirow{3}{*}{$f_{\text{Cut}}$ }&    $T'=T$ (Loss attack)    &        20.6 &                     14.61 &                45.90 &                16.70 &                     2.01 &                   4.6 \\
&$T'=200$           &                32.8 &                     \textbf{33.71} &                58.58 &                20.98 &                     7.89 &                   6.5 \\
& $T'=300$                &                25.4 &                     22.47 &                54.48 &                17.72 &                     4.33 &                   4.7 \\
\multirow{3}{*}{$f_{\text{Rep, Cut}}^1$  } &        $T'=T$   &     19.0 &                      5.62 &                36.57 &                18.74 &                     2.17 &                   0.8 \\
&          $T'=200$  &      \textbf{40.0} &                      3.37 &                58.96 &                20.77 &                     \textbf{8.82} &                   0.9 \\
&      $T'=300$  &           24.8 &                      3.37 &                51.87 &                18.13 &                     4.33 &                   0.7 \\
\multirow{3}{*}{$f_{\text{Rep, Cut}}^2$}            &     $T'=T$ &            19.2 &                      5.62 &                33.58 &                19.35 &                     1.70 &                   4.7 \\
&       $T'=200$ &          33.8 &                      3.37 &                57.84 &                20.57 &                     \textbf{8.82}&                   6.4 \\
&      $T'=300$ &           25.0 &                      3.37 &                50.00 &                18.13 &                     4.80 &                   4.8 \\\hline
\multirow{3}{*}{ $f_{\text{Cal}}$ }     &  $T'=T$       &       16.2 &                      6.74 &                69.78 &                20.37 &                     3.56 &                   6.5 \\
&    $T'=200$     &        28.4 &                      4.49 &                70.90 &                14.26 &                     6.50 &                   6.5 \\
&            $T'=300$   &  20.8 &                      4.49 &                72.76 &                13.24 &                     2.48 &                  \textbf{ 7.5} \\
\multirow{3}{*}{$f_{\text{Rep, Cal}}^1$} &  $T'=T$ &             16.6 &                      4.49 &                67.16 &                19.96 &                     3.56 &                   0.8 \\
&  $T'=200$      &     35.2 &                      4.49 &                66.79 &                16.70 &                     5.26 &                   0.9 \\
&  $T'=300$  &                23.4 &                      3.37 &                67.91 &                14.05 &                     4.80 &                   0.8 \\
\multirow{3}{*}{$f_{\text{Rep, Cal}}^2$} &      $T'=T$ &           15.8 &                      6.74 &                69.40 &                22.00 &                     3.56 &                   6.9\\
&  $T'=200$      &                28.4 &                      5.62 &                66.79 &                14.46 &                     6.35 &                   5.9 \\
&  $T'=300$  &                 21.8 &                      3.37 &                68.66 &                13.65 &                     2.94 &                   7.4 \\\hline
\multirow{3}{*}{$f_{\text{PPL}}$}     &    $T'=T$ &            20.6 &                     14.61 &                45.90 &                16.70 &                     2.01 &                   4.6 \\
&$T'=200$           &                32.8 &                     33.71 &                58.58 &                20.98 &                     7.89 &                   6.5 \\
&$T'=300$       &                25.4 &                     22.47 &                54.48 &                17.72 &                     4.33 &                   4.7 \\
\multirow{3}{*}{$f_{\text{Rep,PPL}}^1$}             &   $T'=T$ &                20.8 &                      2.25 &                36.19 &                16.90 &                     1.70 &                   0.8 \\
&$T'=200$           &                 35.4 &                      0.00 &                58.21 &                19.76 &                     8.67 &                   0.9 \\
&$T'=300$           &                25.8 &                      1.12 &                50.37 &                15.07 &                     5.42 &                   0.5 \\
\multirow{3}{*}{$f_{\text{Rep, PPL}}^2$}  &      $T'=T$         &                20.6 &                      2.25 &                33.58 &                18.33 &                     1.70 &                   4.7 \\
&         $T'=200$ &       32.6 &                      1.12 &                58.21 &                19.76 &                     8.51 &                   6.4 \\
 &    $T'=300$ &                 26.0 &                      2.25 &                49.63 &                17.11 &                     4.95 &                   4.8 \\
\bottomrule
\end{tabular}}
\end{table*}

\begin{table*}[]
\centering
\caption{Performance of using each individual signal. We show the TPR at 1\% FPR for the Pythia-deduped model (2.8B). }
\label{tab:individual_signals_pythia_2.8_2}
\resizebox{0.8\textwidth}{!}{%
\begin{tabular}{llrrrrrr}
\toprule
Feature& Configurations&  Arxiv &  Mathematics&  GitHub &  PubMed &  HackerNews &  Pile-CC\\
\midrule

\multirow{3}{*}{$f_{\text{Cal, PPL}}$} &  $T'=T$    &                16.6 &                      6.74 &                72.39 &                21.18 &                     2.48 &                   6.7 \\
&  $T'=200$       &                34.8 &                      4.49 &                \textbf{73.88} &                23.22 &                     6.04 &                   5.9 \\
&  $T'=300$        &                25.2 &                      4.49 &               \textbf{ 73.88} &                19.35 &                     4.18 &                  \textbf{ 7.5} \\
\multirow{3}{*}{$f_{\text{Rep, Cal, PPL}}^1$}   &     $T'=T$ &            16.6 &                      4.49 &                57.46 &                19.14 &                     2.79 &                   0.8 \\
&  $T'=200$       &                38.2 &                      2.25 &                67.91 &                20.77 &                     5.57 &                   0.9 \\
&  $T'=300$        &                  27.6 &                      2.25 &                65.30 &                15.68 &                     4.02 &                   0.5 \\
\multirow{3}{*}{$f_{\text{Rep, Cal, PPL}}^2$}   &     $T'=T$ &             16.0 &                      4.49 &                57.84 &                23.01 &                     2.48 &                   6.2 \\
&  $T'=200$       &                36.4 &                      2.25 &                67.54 &                21.59 &                     5.42 &                   6.6 \\
&  $T'=300$        &                 27.0 &                      2.25 &                65.30 &                21.18 &                     3.87 &                   7.7 \\\hline
\multirow{3}{*}{$f_{CB}$}         &  $T'=200, \tau = 1$        &     30.4 &                     11.24 &                59.70 &                 5.50 &                     5.11 &                   4.9 \\
&  $T'=200, \tau = 2$         &                31.0 &                      4.49 &                61.94 &                14.46 &                     5.88 &                   6.1 \\
&  $T'=200, \tau = 3$          &                27.6 &                     12.36 &                63.06 &                17.52 &                     7.12 &                   5.0 \\
\multirow{3}{*}{$f_{\text{Rep, CB}}^1$} &  $T'=200, \tau =1$ &                37.0 &                      6.74 &                57.46 &                 5.70 &                     5.73 &                   0.5 \\
&  $T'=200, \tau =2$     &                33.4 &                      7.87 &                60.07 &                12.63 &                     6.81 &                   0.6 \\
&  $T'=200, \tau =3$      &                31.4 &                     20.22 &                60.45 &                18.74 &                     7.43 &                   0.6 \\
\multirow{3}{*}{$f_{\text{Rep, CB}}^2$}     &   $T'=200, \tau =1$  &            32.8 &                     11.24 &                56.72 &                 5.70 &                     5.11 &                   4.6 \\
&     $T'=200, \tau =2$ &            30.8 &                     16.85 &                60.82 &                15.07 &                     4.95 &                   6.1 \\
&     $T'=200, \tau =3$ &              26.0 &                     16.85 &                62.69 &                19.76 &                     7.59 &                   4.5 \\\hline
\multirow{3}{*}{$f_{CBM}$}          &  $T'=T$  &          7.6 &                      0.00 &                 0.00 &                 0.81 &                     3.25 &                   2.4 \\
&       $T'=200$ &             12.4 &                      4.49 &                58.58 &                 3.26 &                     4.18 &                   4.7 \\
&       $T'=300$          &                14.2 &                      2.25 &                39.18 &                 3.05 &                     3.72 &                   2.5 \\
\multirow{3}{*}{$f_{\text{Rep, CBM}}^1$}     &   $T'=T$      &                 1.2 &                     10.11 &                26.49 &                 1.02 &                     1.70 &                   0.2 \\
&   $T'=200$      &               18.2 &                      5.62 &                51.49 &                 2.04 &                     1.70 &                   0.5 \\
&   $T'=300$      &                  10.6 &                     12.36 &                40.67 &                 1.22 &                     0.62 &                   0.5 \\
\multirow{3}{*}{$f_{\text{Rep, CBM}}^2$} & $T'=T$      &                 1.2 &                      8.99 &                14.93 &                 0.41 &                     1.39 &                   1.6 \\
& $T'=200$     &                 7.8 &                      5.62 &                29.85 &                 0.61 &                     2.01 &                   2.1 \\
& $T'=300$     &                 4.2 &                      7.87 &                24.25 &                 1.02 &                     0.62 &                   1.7 \\ \hline
\multirow{3}{*}{$f_{\text{LZ}}$}      &   Number of bins: 3      &        5.8 &                      0.00 &                11.57 &                 2.04 &                     2.94 &                   2.8 \\
&Number of bins: 4         &                 8.8 &                      1.12 &                19.03 &                 3.87 &                     2.79 &                   4.2 \\
&Number of bins: 5        &                 8.6 &                      1.12 &                24.63 &                 3.67 &                     2.63 &                   4.0 \\
\multirow{3}{*}{$f_{\text{Rep, LZ}}^1$}   &    Number of bins: 3   &           6.4 &                      0.00 &                39.18 &                 1.22 &                     1.24 &                   1.1 \\
&     Number of bins: 4   &          9.4 &                      0.00 &                42.16 &                 3.05 &                     1.08 &                   0.7 \\
&      Number of bins: 5 &           9.6 &                      1.12 &                42.91 &                 4.48 &                     2.01 &                   0.9 \\
\multirow{3}{*}{$f_{\text{Rep, LZ}}^2$}    &        Number of bins: 3 &          7.4 &                      0.00 &                31.34 &                 2.85 &                     1.24 &                   3.2 \\
&     Number of bins: 4 &               9.6 &                      0.00 &                39.93 &                 4.07 &                     3.10 &                   3.6 \\
&         Number of bins: 5 &           6.8 &                      1.12 &                38.06 &                 4.68 &                     1.70 &                   3.7 \\\hline
\multirow{3}{*}{$f_{\text{CBPM}}$}    & $T'=T$     &           4.4 &                      2.25 &                24.25 &                 1.83 &                     2.48 &                   2.8 \\
&     $T'=200$ &            19.2 &                      3.37 &                50.75 &                 2.65 &                     4.64 &                   4.4 \\
&    $T'=300$ &             9.4 &                      1.12 &                29.85 &                 1.63 &                     2.48 &                   3.0 \\
\hline
\multirow{3}{*}{$f_{\text{Slope}}$}               &     $T'=600$     &          32.0 &                      1.12 &                50.00 &                 5.30 &                     3.87 &                   4.9 \\
&      $T'=800$      &        27.8 &                     13.48 &                42.54 &                \textbf{24.03} &                     3.41 &                   4.6 \\
&    $T'=1000$   &             20.4 &                      3.37 &                50.00 &                21.18 &                     2.48 &                   3.7 \\\hline
\multirow{3}{*}{$f_{\text{ApEn}}$}  &     $T'=600$     &      9.2 &                      0.00 &                 4.48 &                 1.02 &                     1.70 &                   1.7 \\
&           $T'=800$    &    8.8 &                      0.00 &                 0.37 &                 1.63 &                     2.01 &                   1.8 \\
&           $T'=1000$ &    10.0 &                      0.00 &                 0.37 &                 1.43 &                     3.25 &                   2.2 \\
\bottomrule
\end{tabular}}
\end{table*}

\subsection{Visualizing Signal Distributions Across Domains}

\begin{figure*}[ht!]
    \centering
    \textbf{Arxiv}\\
    \includegraphics[width=0.8\linewidth]{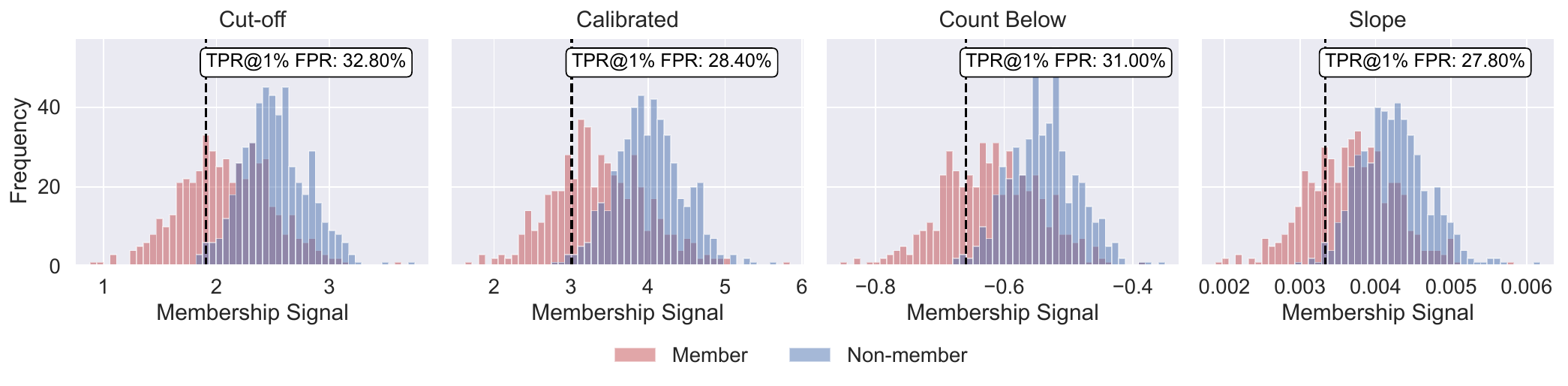}\\
    \textbf{Mathematics}\\
    \includegraphics[width=0.8\linewidth]{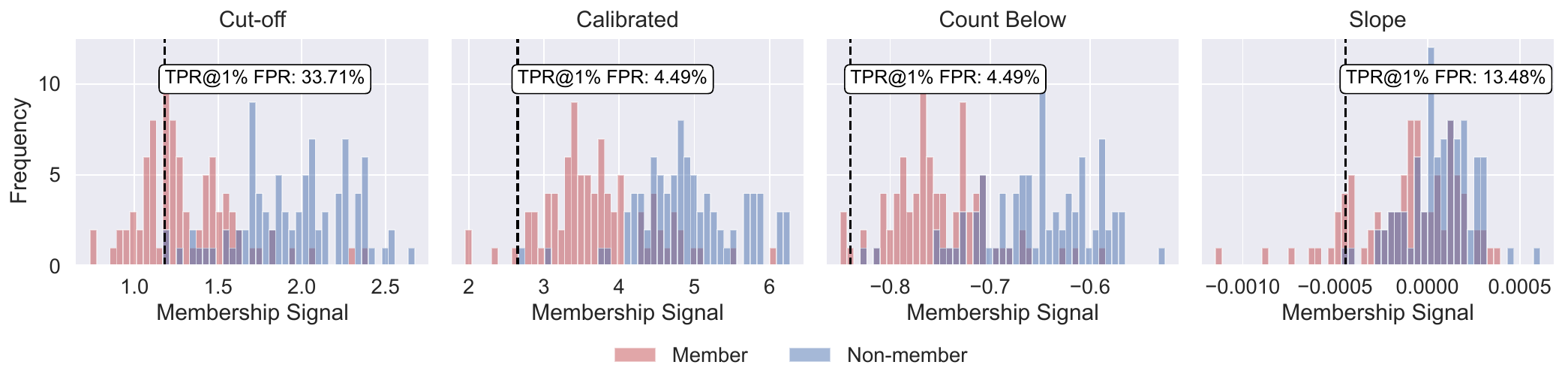}\\
    \textbf{GitHub}\\
    \includegraphics[width=0.8\linewidth]{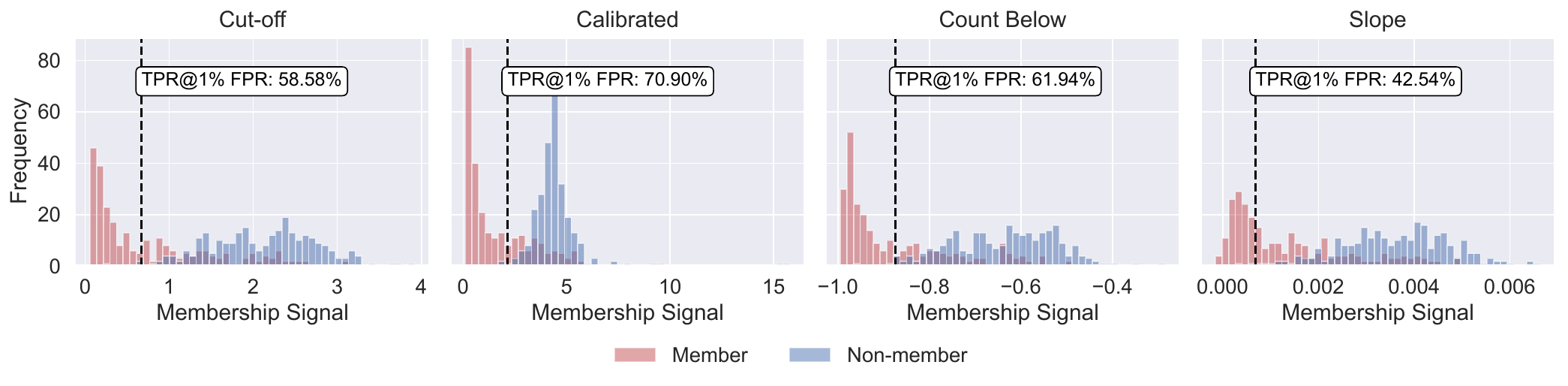}\\
    \textbf{PubMed}\\
    \includegraphics[width=0.8\linewidth]{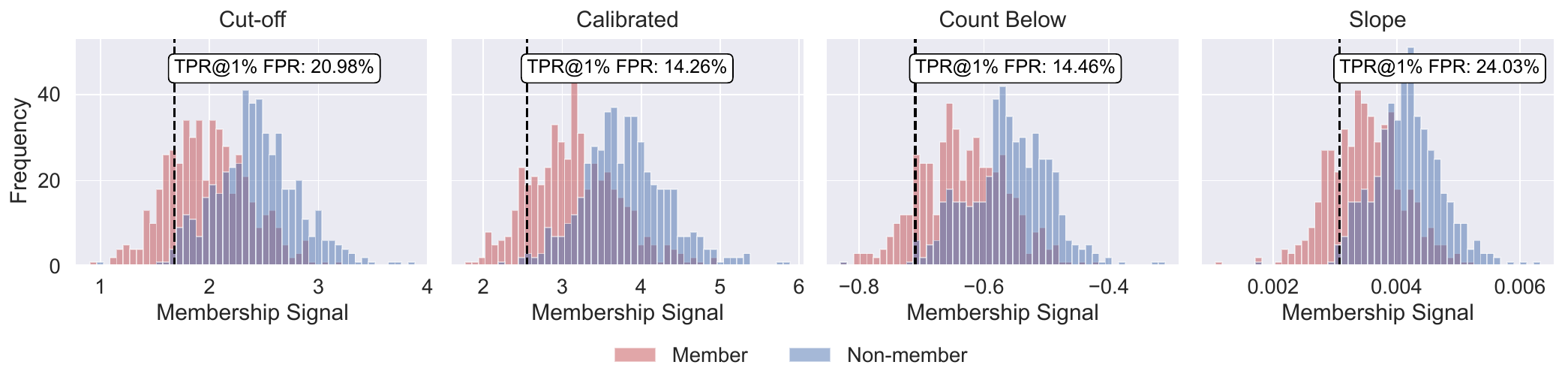}\\
    \textbf{HackerNews}\\
    \includegraphics[width=0.8\linewidth]{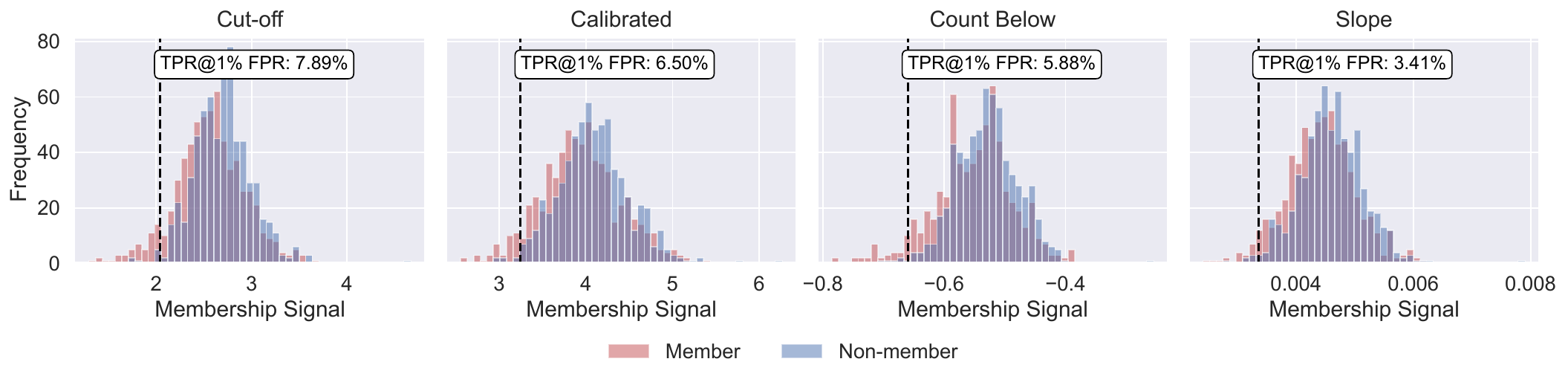}\\
    \textbf{Pile-CC}\\
    \includegraphics[width=0.8\linewidth]{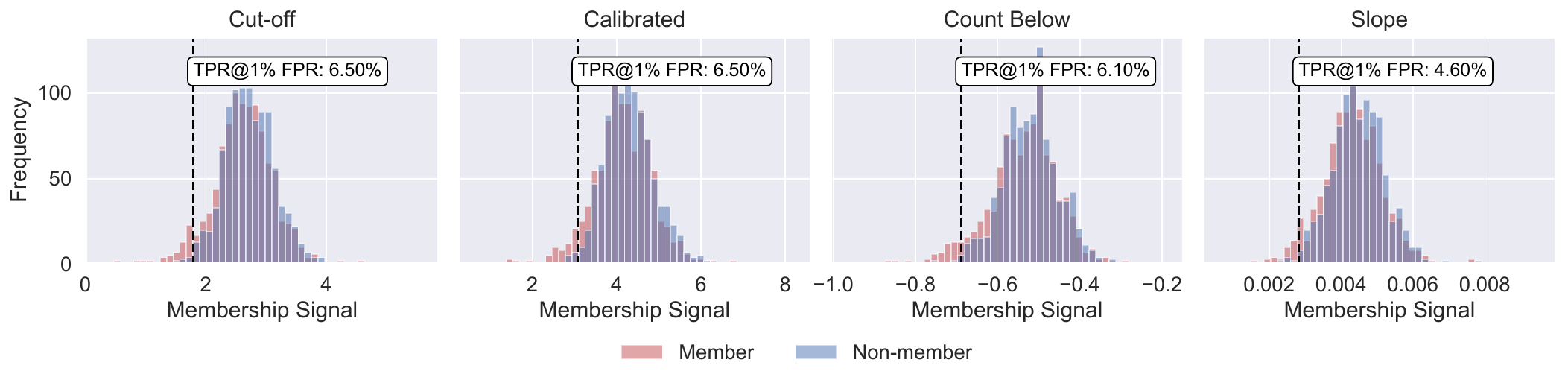}
    \caption{Membership signal distributions for easier domains (Pythia, 2.8B).}
    \label{fig:hist_all_signal_all_domains_easy}
\end{figure*}

Figures~\ref{fig:hist_all_signal_all_domains_easy} illustrates signal distributions for members and non-members across domains, providing visual confirmation that signals effectively distinguish membership, yet vary by domain.

\subsection{Logistic Regression for Signal Combination (Access to Member Data)}\label{appendix:lr_eva}

We detail our LR-based signal combination (introduced in Appendix~\ref{sec:lr_combine}). Specifically, we train a logistic regression model using both member and non-member data to combine individual signals into a unified membership prediction. For training, we sample \(\alpha\%\) of both member (train set) and non-member (test set) data to form the attack dataset. The remaining data serves as the evaluation set. Results in Table~\ref{tab:mia_7_0.2_small_lr} confirm LR-based \name significantly enhances performance.

\begin{table*}[t!]
    \centering
    \caption{Effectiveness of attacks on the Pythia-deduped model with 2.8B. We report the True Positive Rate (TPR) (in \%) at 1\% False Positive Rate (FPR) and the AUC, which quantifies the area under the TPR-FPR curve with FPR ranges from 0 to 1. Higher AUC and TPR indicate better attack performance. The results are averaged over $10$ runs across different random splits of the attack's training and test datasets.    }
    \label{tab:mia_7_0.2_small_lr}
    \resizebox{\textwidth}{!}{%
    \small
    \setlength{\tabcolsep}{2.5pt}
    \begin{tabular}{lcccccccccccccc}
    \toprule
     \multirow{2}{*}{Attack}&\multicolumn{2}{c}{Arxiv} &\multicolumn{2}{c}{Mathematics} & \multicolumn{2}{c}{Github} & \multicolumn{2}{c}{PubMed} & \multicolumn{2}{c}{HackerNews} & \multicolumn{2}{c}{Pile-CC} & \multicolumn{2}{c}{Wikipedia} \\ 
     \cmidrule(lr){2-3}\cmidrule(lr){4-5}\cmidrule(lr){6-7} \cmidrule(lr){8-9} \cmidrule(lr){10-11} \cmidrule(lr){12-13}  \cmidrule(lr){14-15}  
    &  AUC &  TPR&  AUC &  TPR&    AUC &  TPR&  AUC &  TPR &   AUC &  TPR&  AUC &  TPR & AUC &  TPR \\\midrule
Blind~\cite{das2024blind}&0.76 & 0.0&0.93 & 65.95&0.84 & 32.12&0.75 & 0.0&0.53 & 1.94&0.55 & 2.4&0.59 & 0.0\\\hline
\multicolumn{15}{c}{Use Population/Non-member Data} \\\hline
LOSS~\cite{yeom2018privacy} & 0.78 & 14.94 & 0.91 & 12.70 & 0.88 & 39.84 & \textbf{0.79} & 18.20 & 0.60 & 1.06 & 0.55 & 4.77 & \textbf{0.67} & 12.37 \\
Zlib~\cite{carlini2021extracting} & 0.78 & 10.60 & 0.82 & 8.10 & \textbf{0.91} & 46.12 & 0.78 & 14.30 & 0.59 & 2.05 & 0.54 & 5.67 & 0.63 & 9.44 \\
Min-K\%~\cite{shi2023detecting} & 0.75 & 20.11 & \textbf{0.93} & \textbf{46.83} & 0.88 & 40.64 & \textbf{0.79} & 19.45 & 0.58 & 0.84 & 0.54 & 4.56 & 0.66 & 11.53 \\
MIN-K\%++~\cite{zhang2024min} & 0.65 & 5.20 & 0.72 & 17.94 & 0.85 & 31.91 & 0.67 & 10.44 & 0.57 & 1.43 & 0.53 & 2.87 & 0.64 & 10.24 \\
Reference~\cite{carlini2022membership} & 0.71 & 5.86 & 0.42 & 0.00 & 0.73 & 4.68 & 0.63 & 1.22 & 0.57 & 2.63 & 0.58 & 5.93 & 0.68 & 7.36 \\
Neighborhood~\cite{mattern2023membership} & 0.64 & 1.43 & 0.34 & 12.06 & 0.76 & 3.67 & 0.70 & 4.27 & 0.56 & 1.83 & 0.52 & 2.01 & 0.62 & 4.63 \\
\rowcolor{gray!10} \name (Edgington) & \textbf{0.81} & 23.91 & 0.84 & 26.51 & \textbf{0.91} & \textbf{63.30} & \textbf{0.79} & 15.78 & \textbf{0.61} & 4.86 & 0.55 & \textbf{7.39} & 0.66 & 10.26 \\
\rowcolor{gray!10} \name (George) & \textbf{0.81} & \textbf{32.00} & 0.89 & 20.63 & 0.90 & 61.33 & \textbf{0.79} & \textbf{19.94} & \textbf{0.61} & \textbf{5.56} & 0.55 & 6.76 & 0.66 & \textbf{13.56} \\
 \hline
\multicolumn{15}{c}{Use Member and Population/Non-member Data}\\
\hline
    Loss & 0.78 & 15.57 & 0.91 & 16.19 & 0.88 & 40.85 & 0.79 & 19.01 & \textbf{0.60} & 1.41 & 0.55 & 4.86 & 0.67 & 12.76 \\
Zlib & 0.78 & 11.37 & 0.82 & 10.63 & \textbf{0.91} & 47.39 & 0.78 & 14.71 & 0.59 & 2.45 & 0.54 & 5.80 & 0.63 & 9.66 \\
MIN-K\% & 0.75 & 20.80 & 0.93 & 52.06 & 0.88 & 41.54 & 0.79 & 20.03 & 0.58 & 0.97 & 0.54 & 4.74 & 0.66 & 12.00 \\
MIN-K\%++ & 0.65 & 5.60 & 0.72 & 19.37 & 0.85 & 32.71 & 0.67 & 10.67 & 0.58 & 1.52 & 0.53 & 3.06 & 0.64 & 10.37 \\
Reference & 0.71 & 6.51 & 0.44 & 1.27 & 0.73 & 5.16 & 0.63 & 1.77 & 0.57 & 3.22 & \textbf{0.58} & \textbf{6.30} & \textbf{0.68} & 7.70 \\
Neighborhood & 0.64 & 1.63 & 0.67 & 51.11 & 0.77 & 5.59 & 0.70 & 3.81 & 0.56 & 1.83 & 0.51 & 2.24 & 0.62 & 4.93 \\
\rowcolor{gray!10} \name (LR) & \textbf{0.82} & \textbf{34.77} & 0.93 & 29.68 & \textbf{0.91} & \textbf{72.29} & \textbf{0.83} & \textbf{27.62} & 0.59 & \textbf{4.53} & 0.55 & 4.30 & 0.67 & \textbf{15.23} \\
\rowcolor{gray!10} \name (LR + Group PCA) & 0.81 & 32.89 & \textbf{0.95} & \textbf{72.22} & \textbf{0.91} & 64.57 & 0.82 & 26.72 & \textbf{0.60} & 4.46 & 0.54 & 2.70 & 0.66 & 8.96 \\
  \bottomrule
    \end{tabular}}
    \end{table*}

\paragraph{Impact of Dimensionality Reduction on LR.}
We assess dimensionality reduction (via PCA) to manage redundancy in the signal feature set. Table~\ref{tab:effect_of_pca} shows that reducing each signal group's dimensions to 2 (Group PCA with $c=2$) achieves optimal performance across most domains.

\begin{table*}[t!]
\centering
\caption{Effect of Dimensionality Reduction. We show the TPR (in \%) at 1\% FPR of \name when the attack model (logistic regression) is trained on the raw feature, the features after pre-processed by PCA (reduce to $c=10$ features) and group PCA (all membership signals within a group is reduced to $c=1, 2,$ and $3$ features). 
}
\label{tab:effect_of_pca}
\begin{tabular}{lrrrrr}
\toprule
\multirow{2}{*}{Domain}& \multirow{2}{*}{Raw} & PCA & \multicolumn{3}{c}{Group PCA}\\\cmidrule(lr){3-3}\cmidrule(lr){4-6} 
 &  &  $c=10$ &  $c=1$&      $c=2$ &      $c=3$ \\
\midrule
Arxiv         &  34.77 &    27.40 &          32.89 &  \textbf{36.94 }&  35.37 \\
Mathematics &  29.68 &    49.68 &          \textbf{72.22} &  69.37 &  57.30 \\
Github        &  \textbf{72.29} &    62.87 &          64.57 &  66.49 &  67.39 \\
PubMed &  \textbf{27.62} &    23.14 &          26.72 &  26.72 &  26.25 \\
Hackernews     &   4.53 &     3.60 &           4.46 &  \textbf{ 5.47 }&   4.48 \\
Pile-CC     &   4.30 &     3.73 &           2.70 &   \textbf{5.21 }&   4.64 \\
\bottomrule
\end{tabular}
\end{table*}

\begin{figure*}[ht!]
    \centering
    \includegraphics[width=\textwidth]{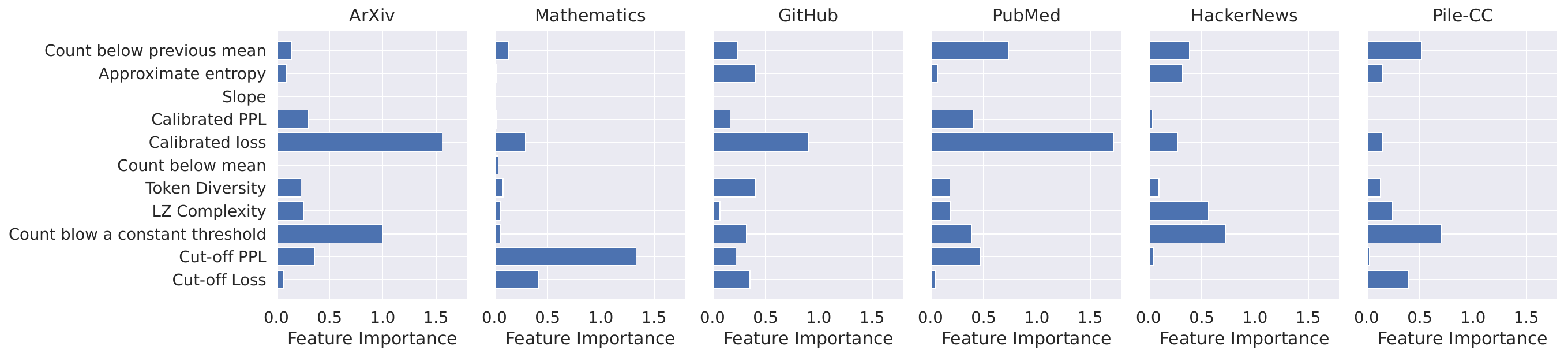}
    \caption{Logistic regression signal importance (Pythia-deduped, 2.8B).}
    \label{fig:feature_importance}
\end{figure*}

\paragraph{Signal Importance Analysis.}
Figure~\ref{fig:feature_importance} visualizes signal importance via LR coefficients. Calibrated loss emerges as dominant in specialized domains (e.g., GitHub), while other signals contribute significantly in varied contexts (e.g., ``count below previous mean'' in Pile-CC), highlighting the advantage of combining diverse signals.

\subsection{Relation Between Model Size, Generalization, and MIA Effectiveness}

Lastly, Table~\ref{tab:effect_model_size} explores relationships between model size, generalization gap (train-test loss difference), and MIA performance. We find no direct correlation with model size; instead, larger generalization gaps strongly correlate with increased memorization and thus higher attack effectiveness, reinforcing generalization's importance for privacy assessment.

\begin{table*}[t!]
\centering
\small
\caption{Model's performance and \name's performance of \name on Pythia-deduped models of different sizes. The ``Gap'' column computes the difference between the model's losses on the training and test data
For \name, we measure its TPR (in \%) at 1\% FPR.}
\label{tab:effect_model_size}
\begin{tabular}{llccccc}
\toprule
\multirow{2}{*}{Dataset} & \multirow{2}{*}{Model} & \multicolumn{3}{c}{Model Performance} &\multicolumn{2}{c}{ Performance of \name}\\\cmidrule(lr){3-5} \cmidrule(lr){6-7}
 && Train &  Test &  Gap &  Edgington & LR+GPR\\ 
\midrule
\multirow{4}{*}{Arxiv} &  
160M      &        2.74 &       3.05 &  0.31 &    23.37 &         24.74 \\
&1.4B      &        2.12 &       2.47 &  0.35 &    25.23 &         31.23 \\
&2.8B      &        1.98 &       2.35 &  0.36 &    25.89 &         32.89 \\
&6.9B     &        1.92 &       2.29 &  0.37 &    28.69 &         33.23 \\
&12B &    1.86 &       2.24 &  0.38 &    28.06 &         36.06 \\\hline
\multirow{4}{*}{Mathematics} 
&160M &    1.46 &       2.14 &  0.68 &    31.11 &         73.97 \\
&1.4B &        1.29 &       1.94 &  0.65 &    11.90 &         71.90 \\
&2.8B &        1.26 &       1.90 &  0.64 &    24.92 &         72.22 \\
&6.9B &        1.25 &       1.89 &  0.64 &    29.05 &         70.79 \\
&12B &        1.24 &       1.87 &  0.63 &    27.62 &         69.84 \\\hline
\multirow{4}{*}{Github}
&160M &        1.41 &       2.49 &  1.08 &    41.81 &         56.91 \\
&1.4B &      0.92 &       1.94 &  1.02 &    54.04 &         57.77 \\
&2.8B &       0.77 &       1.87 &  1.10 &    60.21 &         64.57 \\
&6.9B &        0.77 &       1.77 &  1.01 &    55.32 &         63.72 \\
&12B &        0.71 &       1.72 &  1.01 &    61.38 &         63.78 \\\hline
\multirow{4}{*}{PubMed}
&160M &        2.58 &       3.02 &  0.44 &    21.25 &         30.93 \\
&1.4B &        2.07 &       2.47 &  0.40 &    14.22 &         26.22 \\
&2.8B &        1.97 &       2.36 &  0.39 &    13.14 &         26.72 \\
&6.9B &        1.91 &       2.30 &  0.38 &    11.89 &         24.53 \\
&12B &        1.87 &       2.25 &  0.38 &    11.77 &         21.28\\\hline
\multirow{4}{*}{HackerNews}
&160M &       3.21 &       3.30 &  0.09 &     2.78 &          4.26 \\
&1.4B &        2.60 &       2.70 &  0.10 &     4.99 &          4.55 \\
&2.8B &         2.52 &       2.63 &  0.11 &     4.28 &          4.46 \\
&6.9B &        2.40 &       2.51 &  0.11 &     6.45 &          5.12 \\
&12B &        2.33 &       2.45 &  0.12 &     6.95 &          5.74 \\\hline
\multirow{4}{*}{Pile-CC}
&160M &        3.31 &       3.38 &  0.07 &     4.67 &          1.40 \\
&1.4B &         2.66 &       2.76 &  0.10 &     6.03 &          2.74 \\
&2.8B &       2.58 &       2.68 &  0.10 &     6.94 &          2.70 \\
&6.9B &        2.43 &       2.56 &  0.13 &    10.01 &          4.31 \\
&12B &        2.36 &       2.51 &  0.15 &    10.66 &          4.89\\
\bottomrule
\end{tabular}
\end{table*}

\subsection{Impact of averaging the loss over a small window}
\citet{carlini2021extracting} proposed computing the perplexity over a small sliding window instead of averaging the loss over all tokens. Table~\ref{tab:sliding_window} reports the True Positive Rate (TPR) at 1\% False Positive Rate (FPR) on the Pythia-2.8B model. CAMIA achieves substantially higher TPR than the sliding window approach. The latter computes perplexity only on consecutive sequences of $K$ tokens, which does not necessarily align with where membership signals are strongest. By contrast, CAMIA explicitly identifies and leverages tokens that are most informative given their contextual uncertainty, thereby producing more effective membership inference.

\begin{table*}[t!]
\centering
\small
\begin{tabular}{c|ccccc}
\toprule
K   & Arxiv (\%) & Github (\%) & Pubmed (\%) & Pile-CC (\%) & HackerNews (\%) \\
\midrule
CAMIA & 32.00 & 63.30 & 19.94 & 7.39 & 5.56 \\
\midrule
1   & 1.71  & 13.83 & 10.17 & 2.29 & 2.43 \\
10  & 6.86  & 26.06 & 5.52  & 2.00 & 2.43 \\
20  & 4.29  & 34.57 & 6.40  & 1.43 & 0.00 \\
30  & 11.14 & 25.53 & 5.52  & 2.71 & 0.44 \\
40  & 7.71  & 25.00 & 2.91  & 2.71 & 0.00 \\
50  & 10.00 & 31.91 & 6.69  & 4.14 & 1.99 \\
60  & 9.43  & 27.13 & 4.07  & 3.86 & 2.65 \\
70  & 7.14  & 31.91 & 6.10  & 2.43 & 2.21 \\
80  & 9.14  & 27.66 & 5.23  & 3.86 & 2.87 \\
90  & 12.29 & 26.60 & 4.36  & 3.86 & 1.32 \\
100 & 20.00 & 31.91 & 5.81  & 3.29 & 2.21 \\
110 & 10.86 & 31.91 & 4.65  & 4.14 & 1.77 \\
120 & 10.86 & 37.23 & 4.07  & 4.57 & 1.10 \\
130 & 17.71 & 34.04 & 0.29  & 4.29 & 2.43 \\
140 & 16.29 & 33.51 & 5.52  & 5.00 & 4.64 \\
150 & 24.86 & 33.51 & 11.63 & 3.57 & 1.10 \\
All & 14.94 & 39.84 & 18.20 & 4.77 & 1.06 \\
\bottomrule
\end{tabular}
\caption{TPR at 1\% FPR for Pythia-2.8B under the sliding window method of \citet{carlini2021extracting} (window size $K$) compared to CAMIA.}
\label{tab:sliding_window}
\end{table*}

\end{document}